\documentclass[11pt]{article}

\usepackage[final]{acl}
\usepackage{float}
\usepackage{times}
\usepackage{latexsym}
\usepackage[table]{xcolor}
\usepackage{colortbl}
\usepackage{todonotes}
\usepackage{amsmath}
\usepackage{booktabs}
\usepackage{url} 
\usepackage{hyperref} 
\usepackage{listings}
\usepackage{longtable}
\usepackage{tikz}
\usetikzlibrary{positioning,arrows.meta,fit}
\usetikzlibrary{calc} 

\usepackage[T1]{fontenc}

\usepackage[utf8]{inputenc}

\usepackage{microtype}

\usepackage{inconsolata}
\usepackage{twemojis}

\usepackage{graphicx}
\usepackage{newunicodechar}
\newunicodechar{−}{\ensuremath{-}} 
\newunicodechar{⁻}{\ensuremath{^{-}}} 

\title{AtlasNLP \texttwemoji{globe with meridians}: A Country-Aware Atlas of Dataset Representation in NLP}

\author{
 \textbf{Joan Nwatu\textsuperscript{1}},
 \textbf{Tsedeniya Solomon Amare\textsuperscript{1}},
 \textbf{Longju Bai\textsuperscript{1}},
 \textbf{Bontu Fufa Balcha\textsuperscript{2}},
\\
 \textbf{Zayd Bashir \textsuperscript{3}},
 \textbf{Angana Borah\textsuperscript{1}},
 \textbf{Zara Burzo\textsuperscript{4}},
 \textbf{Yubin Choi \textsuperscript{1}},
\\
 \textbf{Naihao Deng\textsuperscript{1}},
 \textbf{Samika Gupta\textsuperscript{1}},
 \textbf{Michel Faloughi\textsuperscript{5}},
 \textbf{Claude Kwizera\textsuperscript{6}},
\\
 \textbf{Ziqiao Ma\textsuperscript{1}},
 \textbf{ Cynthia Yacel Fuertes Panizo\textsuperscript{1}},
 \textbf{Ellie Seehorn\textsuperscript{1}},
 \textbf{Hui Shen\textsuperscript{1}},
\\
 \textbf{Jiayi Tang\textsuperscript{1}},
 \textbf{Zesen Zhao\textsuperscript{1}},
 \textbf{Boyuan Zheng\textsuperscript{1}},
 \textbf{Rada Mihalcea\textsuperscript{1}}
\\
\\
 \textsuperscript{1}University of Michigan, USA,
 \textsuperscript{2}Addis Ababa University, Ethiopia,
 \\
 \textsuperscript{3}San Jose State University, USA,
 \textsuperscript{4}Skyline High School, USA,
 \\
 \textsuperscript{5}University of Pennsylvania, USA,
  \textsuperscript{6}Carnegie Mellon University, Rwanda
\\
\texttt{\{\href{mailto:jnwatu@umich.edu}{jnwatu}, \href{mailto:mihalcea@umich.edu}{mihalcea}\}@umich.edu}
}

\begin{document}
\maketitle
\begin{abstract}

Understanding which countries are represented in NLP datasets is essential for identifying gaps, targeting data collection, measuring progress, and informing AI policy. However, geographic metadata is very rarely available, and country-level representation is often hidden behind broad language-level claims. We introduce \textbf{AtlasNLP}, a country-aware atlas of over 13,000 NLP dataset records across normalized NLP task categories, tracking both the populations represented and where datasets are produced. AtlasNLP includes  \textbf{AtlasNLP-Gold}, a human-curated reference set, and \textbf{AtlasNLP-Core}, an ACL-derived large-scale collection. Using this resource, we show that (1) dataset coverage is highly uneven across countries and tasks; (2) dataset production and representation are geographically asymmetric; and (3) language coverage does not imply geographic representation. These findings reveal blind spots in current dataset documentation practices and motivate more explicit geographic metadata for country-aware NLP evaluation. We release AtlasNLP at \href{https://lit.eecs.umich.edu/AtlasNLP/index.html}{the AtlasNLP project site}.

\end{abstract}

\section{Introduction}

Efforts to improve representation in NLP increasingly emphasize broader dataset coverage across languages, regions, and populations \cite{burchell2024improving,bender2018data}. Yet it remains difficult to determine which countries and populations are represented, where gaps remain, and whether coverage is improving \cite{yu2022beyond,ranathunga2022some}. Most dataset collections are organized by language, task, or benchmark, providing limited visibility into the geography of dataset representation.

\begin{figure}[h]
  \includegraphics[width=\columnwidth,trim=0 55 0 55,clip]{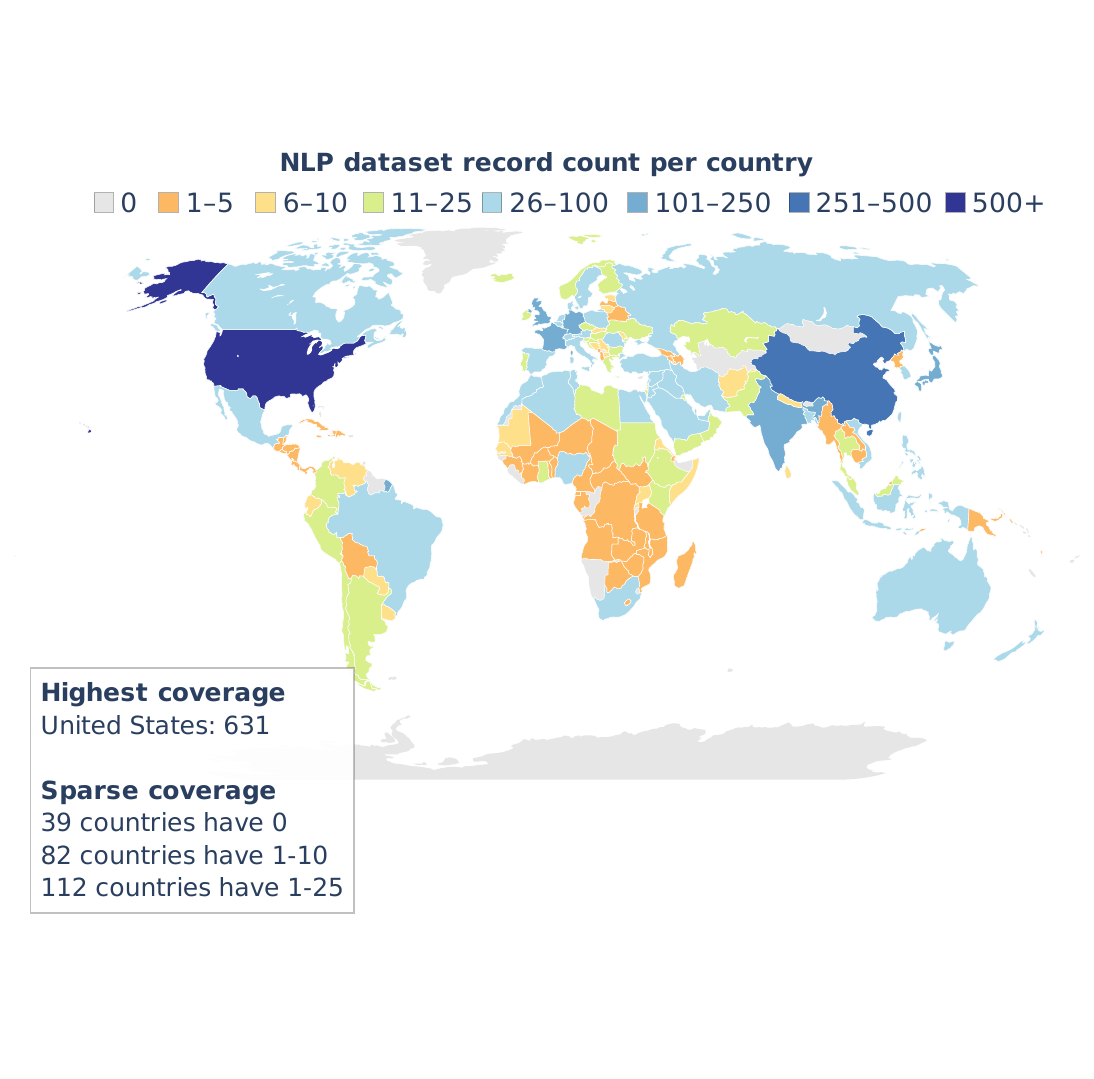}
    \caption{Global distribution of NLP dataset coverage in AtlasNLP-Core. Countries are bucketed by represented dataset record count, showing a highly uneven distribution with a small number of high-coverage countries and many sparsely represented regions.}
  \label{fig:world_map}
\end{figure}

This lack of visibility matters because dataset coverage is not globally uniform. As Figure~\ref{fig:world_map} shows, a small number of countries account for a large share of represented dataset records, while many others remain sparsely represented or absent. Coverage also varies across tasks: a country may have datasets for machine translation or sentiment analysis, but little or no coverage for question answering, reasoning, safety, or multimodal understanding. These gaps limit our ability to evaluate NLP systems consistently across countries and populations and to identify where additional dataset development is most needed \cite{alabi2025charting}. Importantly, this also makes it difficult for local governments to plan and assess AI initiatives, since they cannot easily determine what resources are available for their countries or where new investment is most needed.

A central reason these patterns are difficult to measure is that geographic metadata is rarely documented explicitly \cite{faisal2022dataset}. Dataset documentation commonly records language and task, but often omits which countries or populations are represented \cite{bender2018data, faisal2022dataset}. Language metadata provides an important but incomplete signal: many languages span multiple countries, and many countries contain multiple linguistic communities \cite{aji2022one,hershcovich2022challenges, tonneau2024languages}. For example, Spanish is spoken across Spain and much of Latin America, yet datasets labeled as Spanish are often representative only for a subset of these countries.  As a result, language-level views can make geographic representation appear broader than the underlying country-level evidence supports, especially when multilingual datasets are treated as evidence of broad population coverage. 

To address these limitations, we introduce \textbf{AtlasNLP}, a country-aware atlas of over 13,000 NLP dataset records covering 31 normalized NLP task categories such as machine translation and question answering. Alongside task and language metadata, AtlasNLP records both \textit{represented countries}, i.e., the countries or populations represented in a dataset; and \textit{producer countries}, i.e., the institutional locations where datasets are produced. The resource consists of two complementary components:  \textbf{AtlasNLP-Gold}, which is a human-curated reference set; and \textbf{AtlasNLP-Core}, an ACL-derived large-scale collection constructed through automated extraction. Using AtlasNLP-Gold, we validate AtlasNLP-Core, and broaden AtlasNLP's coverage of underrepresented regions. Using AtlasNLP-Core, we analyze patterns of geographic representation, task coverage, and dataset production within the ACL literature.

Our contributions are as follows: \textbf{(1)} we release AtlasNLP-Core and AtlasNLP-Gold as country-aware resources for studying NLP dataset geography; \textbf{(2)} we show that dataset coverage is highly uneven across countries and tasks, with 79.2\% of country-task pairs containing no  dataset records with documented country representation; \textbf{(3)} we distinguish {\it representation} from {\it production}, showing that many countries are represented primarily through datasets produced outside the country; \textbf{(4)} we show that dataset availability is positively associated with broader research infrastructure, including national income and university capacity; and \textbf{(5)} we show that language coverage does not imply geographic coverage, motivating more explicit country-level dataset documentation.

\section{Related Work}
\paragraph{Geographic Representation of NLP Datasets.} 
NLP datasets, benchmarks, and model ecosystems are geographically uneven \cite{bender2021dangers,blodgett2020language}. AI systems often perform better for populations that are better represented in training and evaluation data \cite{mihalcea2025ai,nwatu2026culture}, and model behavior can vary across populations even when tasks appear similar \cite{hershcovich2022challenges,naous2024having}. These disparities have been documented across modalities, including vision-language models \cite{shankar2017no,de2019does,nwatu2023bridging,nwatu2025uplifting} and speech technologies \cite{elmadany2025voice}.

Prior work has shown that representation is not only a matter of dataset quantity: languages, regions, and populations differ in the tasks, resources, and technologies available to them \cite{joshi2020state,blasi2022systematic,yu2022beyond}. However, most studies examine geographic or population bias within individual datasets, models, tasks, languages, or regional settings \cite{hershcovich2022challenges,naous2024having,alabi2025charting, faisal2022dataset,yu2022beyond, longpre2025bridging}, but do not provide a unified country-level view of which populations are represented in NLP datasets across regions, who produces those datasets, and how task coverage varies geographically. AtlasNLP addresses this gap by enabling systematic analysis of NLP dataset representation, production, and task coverage across countries.

\paragraph{Multilingual and Cross-Lingual Evaluation.} 
NLP evaluation is commonly organized around language. Benchmarks such as XTREME \cite{hu2020xtreme}, XGLUE \cite{liang2020xglue}, and FLORES \cite{goyal2022flores} expanded evaluations beyond English, while multilingual surveys document persistent disparities across low-resource and non-Western languages \cite{joshi2020state,blasi2022systematic,ranathunga2022some,joshi2025natural}. However, language is an incomplete proxy for geographic and cultural representation because many languages span multiple countries, and many countries contain multiple linguistic communities \cite{hershcovich2022challenges,blodgett2020language, tonneau2024languages, liu2025culturally, pawar2025survey}. AtlasNLP  adds country-level structure alongside language metadata, making it possible to study variation in representation, task coverage, and production within and across languages.

\paragraph{Dataset Documentation, Platforms, and Metadata.} 
Dataset documentation work emphasizes recording dataset origins, collection procedures, intended uses, and potential biases \cite{gebru2021datasheets,bender2018data, longpre2024data}. Dataset platforms such as Hugging Face and  publication repositories like ACL Anthology have also made resources easier to discover and reuse \cite{lhoest2021datasets}. Yet geographic metadata remains underdeveloped: language and task metadata are often available, while country-level information is frequently incomplete, implicit, or absent. This makes it difficult to distinguish \textit{represented geography}, in which populations are represented, from \textit{producer geography}, in which datasets are institutionally created \cite{santy-etal-2023-nlpositionality}. AtlasNLP addresses this gap by introducing structured country-aware metadata for over 13,000 NLP dataset records.

\section{Methodology}

Constructing a country-aware view of the NLP dataset ecosystem requires more than extracting existing metadata. Geographic information is often implicit, inconsistently documented, or absent, and language alone is frequently insufficient for determining which populations are represented. We develop AtlasNLP through three stages: (i) designing a structured annotation framework and human-curated reference collection, (ii) scaling the framework over ACL Anthology papers, and (iii) auditing and refining the extracted dataset records to produce a high-confidence dataset collection.

\subsection{Annotation Framework and Schema Design}

We design a structured annotation framework that records: (i) represented countries, (ii) producer countries, (iii) task category, (iv) language, (v) modality and licensing, and (vi) attribution method. Task categories are derived from recent ACL and EMNLP thematic areas and normalized into 30 categories. AtlasNLP-Gold additionally includes \textit{Raw Corpus} for unlabeled or minimally structured text collections.

We use a standardized set of 197 geopolitical entities, consisting of the 193 UN member states and two UN observer states, together with Taiwan and Kosovo. A central design choice is the distinction between \textit{represented country}, the country or population represented in a dataset, and \textit{producer country}, the institutional location of its creators. This allows us to analyze dataset representation separately from dataset production.

Represented-country evidence is classified as \textit{explicit}, \textit{inferred}, or \textit{unattributed}. Explicit attribution requires direct evidence connecting dataset content, participants, data sources, or collection to a country. Inferred attribution captures plausible geographic relevance supported by indirect evidence, such as a geographically specific language variety, community, or data source. Language alone is not sufficient for explicit country attribution. Primary analyses use explicit attribution, while key analyses are repeated with explicit+inferred attribution as a sensitivity analysis. 

\subsection{Human Curation}

We construct a human-curated collection to ground the annotation framework, improve coverage of underrepresented regions, and provide a reference set for validation. Contributors come from diverse geographic backgrounds, including Nigeria, China, Peru, the United States, Romania, South Korea, and Ethiopia.

Contributors start by selecting the countries they are familiar with and collect associated datasets from ACL Anthology, Hugging Face, Papers With Code, GitHub repositories, institutional archives, and regional dataset collections. To encourage balanced coverage, countries are grouped into high-, medium-, and low-resource categories, and contributors select country sets spanning multiple regions. Contributors are encouraged to identify up to five datasets per country, with particular attention to countries with limited NLP data representation. Each dataset is annotated using the shared schema, including dataset name, task category, language, modality, licensing, country attribution, and provenance.

The curation process has two stages. In the \textit{compilation} stage, contributors independently collect and annotate datasets for their assigned countries. In the \textit{cross-validation} stage, assignments are shuffled and redistributed for secondary review. Reviewers verify country attribution, task categorization, metadata consistency, duplicates, and provenance, reducing annotator bias and improving consistency.
The resulting collection, \textbf{AtlasNLP-Gold}, contains 1,480 human-curated entries and serves as a high-quality reference set for evaluating automated extraction. Additional details on Gold construction and review are provided in Appendix~\ref{app:gold}.

\subsection{Automated Dataset Expansion}

To scale beyond manual coverage, we apply the annotation framework to ACL Anthology\footnote{https://aclanthology.org/} publications. We begin with 119,963 ACL records spanning 1952--2025. Papers with abstracts are evaluated using ModernBERT-base-NLI \cite{sileo_2024_tasksource} against the hypothesis {\it ``This paper proposes a dataset.''} We retain papers with entailment scores greater than 0.5, yielding 20,277 candidate dataset papers.

Candidate papers are processed using a staged extraction pipeline combining rule-based dataset-role checks with schema-constrained GPT-4o-mini extraction from full paper text. The pipeline extracts dataset identity, task, language, licensing, dataset role, and geographic evidence, together with page- and quote-level provenance. This produces 18,035 successful initial paper-level extractions.

A subsequent 400-record human audit reveals recurring failure modes, particularly reuse of existing datasets and country assignments based on affiliation, language, or incidental mentions rather than dataset provenance. We use these findings to tighten the dataset-role and country-attribution criteria and re-evaluate the full extracted collection. The post-audit re-evaluation used GPT-5.4-mini for evidence extraction and verification. Dataset records are retained when the paper introduces a dataset, materially extends an existing dataset, or constructs a new dataset compilation as a primary contribution. The resulting \textbf{AtlasNLP-Core contains 13,462 primary dataset-contribution records}. Full extraction, exclusion, and audit details are reported in Appendix~\ref{app:audit}.

\subsection{Validation and Quality Control}

We evaluate the final pipeline using three complementary human-reviewed checks.

First, we compare post-audit AtlasNLP-Gold against the final AtlasNLP-Core. Matching by ACL paper and dataset identity yields 225 high-confidence aligned dataset records. Among matched Gold dataset records for which AtlasNLP-Core assigns an explicit country, 97.7\% share at least one country with the Gold reference set, and 94.7\% of individual country assignments made by Core are supported by Gold. Exact task agreement is 79.9\%. When inferred country evidence is included on both Gold and Core, country-assignment precision remains similar at 94.0\%.

Second, the 400-record audit is used to verify that the failure modes identified in the initial extraction are substantially reduced after revision. Of 187 dataset records whose country attribution was judged incorrect, 177 (94.7\%) no longer remain in the explicit geographic layer. Among the 98 eligible cases for which the human review finds no supported country, 97 are correctly left unattributed. Detailed audit results and recall-oriented analyses are reported in Appendix~\ref{app:audit}.

Finally, because producer geography is derived from author affiliations and is conceptually distinct from represented-country attribution, we evaluate it independently on 100 human-labeled dataset records. Producer-country extraction achieves 96.9\% precision and 97.7\% recall at the country-association level, with the primary producer country recovered in all audited cases.

\begin{figure*}[h]
\centering
  \includegraphics[width=0.97\textwidth]{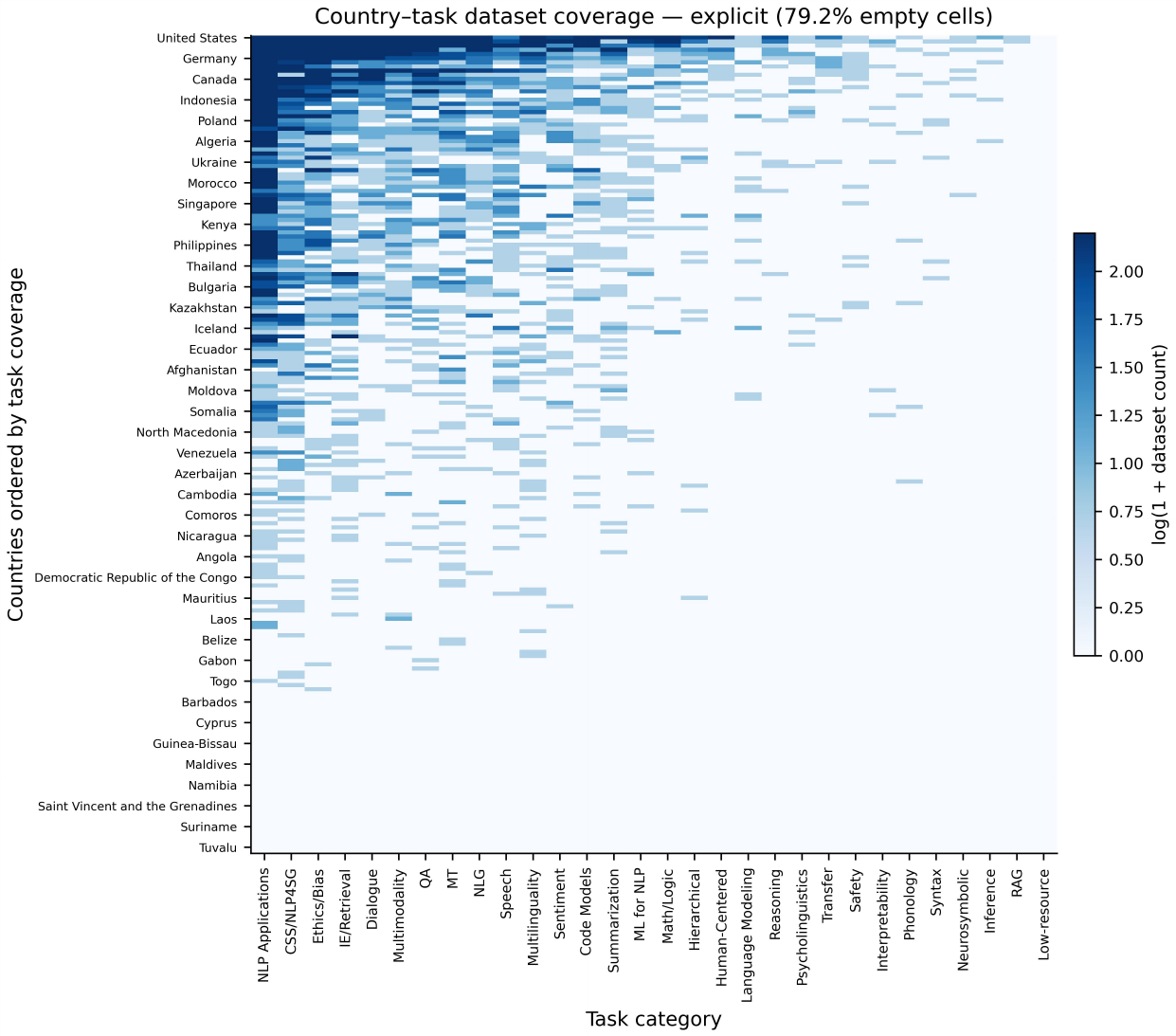}
  \caption{Heatmap of dataset coverage across countries and tasks. Coverage is highly fragmented, with most country-task pairs lacking represented dataset records and even high-coverage countries concentrated in a subset of tasks. Only a subset of country labels is shown for readability, with labels placed at regular intervals (approximately every fifth country). Best viewed in color.}
  \label{fig:country_task_heatmap}
\end{figure*}

\section{AtlasNLP Overview}

AtlasNLP has two complementary components. \textbf{AtlasNLP-Core} contains 13,462 paper-level dataset records corresponding to primary dataset contributions in ACL Anthology publications. We use the term \textit{dataset record} rather than \textit{unique dataset} because related papers may introduce, extend, or compile versions of the same underlying resource, and resolving these relationships into unique dataset entities is not always straightforward. \textbf{AtlasNLP-Gold} contains 1,480 human-curated entries, corresponding to 989 normalized dataset-name groups. We keep the collections separate because they differ in source coverage and metadata completeness: Core provides consistent large-scale metadata, including producer-country information from author affiliations, while Gold supplements the ACL-derived collection with human-curated resources, particularly for underrepresented regions.

Table~\ref{tab:AtlasNLP_stats} summarizes the final AtlasNLP resource. Represented-country evidence is available at two confidence levels. Explicit country attribution is available for 2,447 Core dataset records (18.2\%), while including inferred attribution expands coverage to 3,506  dataset records (26.0\%). Dataset records without sufficiently supported country evidence remain in Core and contribute to analyses that do not require represented-country metadata.  Unless otherwise stated, analyses use AtlasNLP-Core.

\begin{table}[h]
\centering
\scalebox{0.65}{
\begin{tabular}{l r}
\toprule
Statistic & Value \\
\midrule
AtlasNLP-Core  dataset records & 13,462 \\
AtlasNLP-Gold entries / normalized name groups & 1,480 / 989 \\
Core dataset records w/explicit country attribution & 2,447 (18.2\%) \\
Core dataset records w/explicit+inferred attribution & 3,506 (26.0\%) \\
Core countries represented (explicit / +inferred) & 158 / 168 \\
Gold countries represented (explicit / +inferred) & 192 / 197 \\
Core country-record associations (explicit / +inferred) & 4,421 / 6,001 \\
Task categories (Core / AtlasNLP overall) & 30 / 31 \\
Audited languages (Core / AtlasNLP overall) & 1,239 / 1,245 \\
Multilingual dataset records (Core) & 3,063 \\
Producer-country metadata (Core) & 13,224 (98.2\%) \\
\bottomrule
\end{tabular}
}
\caption{Summary statistics for AtlasNLP. AtlasNLP-Core contains vetted primary dataset-contribution records. Represented-country statistics distinguish explicit attribution from the broader explicit+inferred sensitivity layer.}
\label{tab:AtlasNLP_stats}
\end{table}

\section{Dataset Geography Analysis}

\subsection{Dataset Coverage is Highly Uneven Across Countries and Tasks}

We first examine which countries are represented in NLP datasets using
explicit represented-country attribution. Coverage follows a pronounced
long-tail distribution. Across AtlasNLP-Core, 2,447  dataset records have explicit country attribution, corresponding to 4,421 country-record associations
across 158 of 197 countries. The United States, China, India, the United
Kingdom, and Germany account for a large share (36.4\%) of these associations. At the other end of the distribution, 39 countries have no explicitly attributed dataset records, and 121 of 197 countries have ten or fewer represented dataset records.

Coverage is even sparser across countries and tasks. As shown in
Figure~\ref{fig:country_task_heatmap}, nearly four in five country-task
pairs have no explicitly attributed  dataset records (79.2\%). This pattern remains
under explicit+inferred attribution, where three-quarters of pairs are still
empty (75.0\%). Geographic imbalance therefore reflects not only how many represented dataset records each country has, but also which NLP tasks they cover.

Country attribution is also systematically missing. Among unattributed Core
records, 70.7\% include English, compared with 49.9\% of explicitly
attributed dataset records. Because English is globally distributed, language alone
provides little evidence of represented country. We therefore interpret low
or zero coverage as gaps in documented country representation within ACL
dataset contributions, not evidence that no NLP resources exist.

\subsection{Dataset Production and Representation are Asymmetric}

Dataset coverage alone does not show who produces the datasets used to
represent different countries. We therefore compare \textit{represented
country} with \textit{producer country}, derived from the institutional
locations of dataset creators based on author affiliations. Producer geography
captures institutional location rather than researcher identity, reflecting
differences in research capacity, agendas, and environments.

To construct the production-representation analysis, we first restrict the
explicit represented-country layer to records for which producer-country
metadata is available. This reduces the 2,447 explicitly attributed Core
records (4,421 country-record associations) to 2,396 records and 4,305
represented-country associations. Records without recovered producer
geography remain in the coverage and task analyses but cannot contribute to
production-representation ratios. For example, \textit{BizBench}
\citep{krumdick-etal-2024-bizbench} and the Online Tenant Reviews dataset
\citep{haber-waks-2021-classification-geotemporal} have explicit U.S.
representation in AtlasNLP but no recovered producer country.

We then expand each retained represented-country association over all producer
countries associated with that record. Thus, a dataset record representing
$R$ countries and associated with institutions in $P$ producer countries
contributes $R\times P$ producer-representation associations. This expansion
yields 6,533 producer-representation associations from the 2,396 eligible
records and preserves multi-country representation and cross-country
collaboration rather than forcing either side to a single country.

\begin{figure}[h]
  \includegraphics[width=\columnwidth]
  {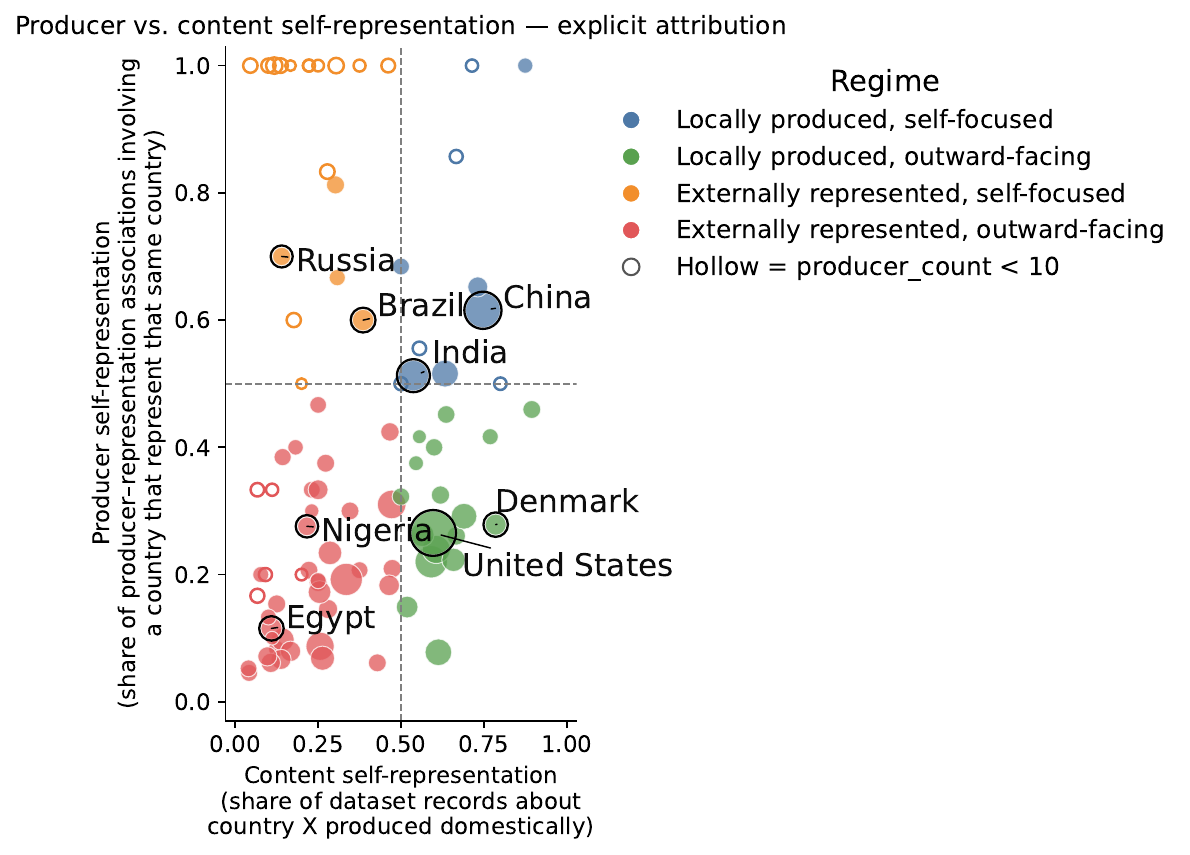}
  \caption{
  Dataset production and representation by country under explicit attribution.
  The x-axis measures the share of datasets about a country produced
  domestically; the y-axis measures the share of producer-representation associations involving a country that represent itself. Dashed lines define four production-representation regimes. Point size reflects dataset volume; hollow points indicate countries with fewer than 10 producer associations.}
  \label{fig:represented_producer}
\end{figure}

Figure~\ref{fig:represented_producer} compares two complementary
measures. The x-axis shows \textit{content self-representation}: among
represented records about a country with producer metadata, the share that
include that same country among their producer countries. The y-axis shows
\textit{producer self-representation}: among expanded
producer-representation associations involving a producer country, the share
whose represented country is that same country. Together, these measures
distinguish countries whose representation is primarily produced locally or
externally, and producers that focus primarily on themselves or on other
countries.

The resulting patterns are strongly asymmetric. Among countries with at least 10 represented records with producer metadata and 10 producer-representation associations, 39 of 62 (62.9\%) have content self-representation below 0.5, meaning that most dataset records representing them were produced by institutions outside the country. This pattern is similar without the minimum-count restriction and under explicit+inferred attribution. 

Figure~\ref{fig:represented_producer} shows the broader distribution, while Table~\ref{tab:producer_examples} reports values for representative countries in each regime. China and India are both locally produced and self-focused, while the United States and Denmark have high domestic representation but produce many datasets about other countries. Nigeria and Egypt are more externally represented and outward-facing. Russia and Brazil show a different pattern: much of their representation is produced externally, while their own dataset production is comparatively self-focused.

\begin{table}[t]
\centering
\setlength{\tabcolsep}{4pt}
\scalebox{0.7}{
\begin{tabular}{lrrrr}
\toprule
\textbf{Country} &
\begin{tabular}[c]{@{}c@{}}\textbf{Content}\\ \textbf{self-rep.}\end{tabular} &
\begin{tabular}[c]{@{}c@{}}\textbf{Producer}\\ \textbf{self-rep.}\end{tabular} &
\begin{tabular}[c]{@{}c@{}}\textbf{Represented}\\ \textbf{records w/prod.}\end{tabular} &
\begin{tabular}[c]{@{}c@{}}\textbf{Producer-rep}\\ \textbf{associations}\end{tabular} \\
\midrule
United States & 0.597 & 0.265 & 620 & 1394 \\
China         & 0.747 & 0.615 & 364 & 442  \\
India         & 0.537 & 0.513 & 227 & 238  \\
Denmark       & 0.786 & 0.278 & 28  & 79   \\
Nigeria       & 0.216 & 0.276 & 37  & 29   \\
Egypt         & 0.109 & 0.115 & 55  & 52   \\
Russia        & 0.140 & 0.700 & 50  & 10   \\
Brazil        & 0.386 & 0.600 & 70  & 45   \\
\bottomrule
\end{tabular}}
\caption{Examples of production-representation asymmetry under explicit
country attribution. Represented-record counts are restricted to records
with producer metadata; producer-side counts are expanded
producer-representation associations. All highlighted countries have at least 10 represented records with producer metadata and 10 producer–representation associations.}
\label{tab:producer_examples}
\end{table}

Production is also geographically concentrated.The United States accounts
for 21.3\% of expanded producer-representation associations, while the ten
largest producer countries account for 61\% of these associations. This concentration persists under explicit+inferred attribution. Together, these results show that geographic representation in NLP depends not only on which countries appear in datasets, but also on where the institutions producing those datasets are located \cite{santy-etal-2023-nlpositionality}.

\subsection{Dataset Availability Tracks Research Infrastructure}

We next examine whether geographic dataset record coverage aligns with broader
research infrastructure. Using World Bank income categories and national
university counts, we compare represented-country coverage with two coarse
indicators of institutional capacity.

\begin{figure}[h]
  \includegraphics[width=0.9\columnwidth]{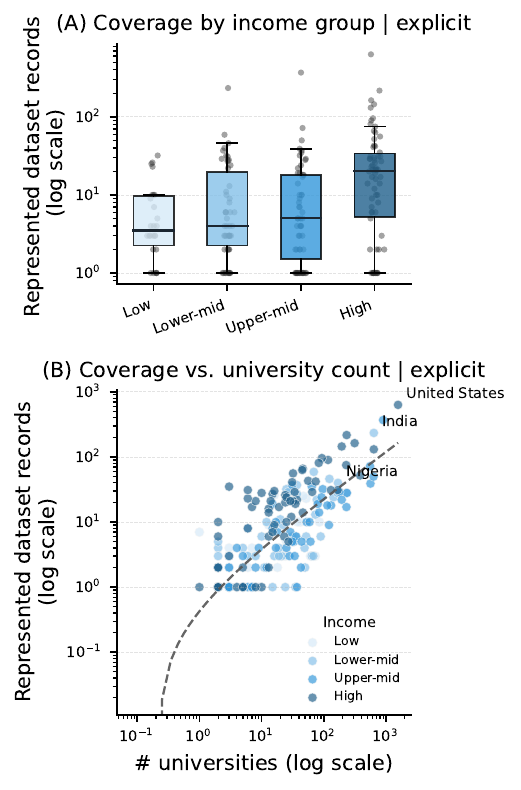}
\caption{
    Relationship between represented-country dataset records and research infrastructure under explicit attribution. (A) Coverage by World Bank income group.
    (B) Coverage increases with national university counts on log-transformed
    values. Color indicates income group. Best viewed in color.}
  \label{fig:infrastructure_alignment}
\end{figure}

Figure~\ref{fig:infrastructure_alignment}(A) shows a strong income gradient:
the median number of represented  dataset records is substantially higher for
high-income countries than for the other income groups.

Figure~\ref{fig:infrastructure_alignment}(B)
similarly shows a positive association between national university counts
and dataset coverage (Pearson $r=.65$; Spearman $\rho=.61$). The relationship
strengthens slightly under explicit+inferred attribution ($r=.70$;
$\rho=.67$).

These patterns show that dataset availability is positively associated with broader research infrastructure. The relationship is not
deterministic, however: countries with similar levels of institutional
capacity can still differ substantially in coverage, reflecting the roles
of language, publication ecosystems, and regional research priorities.

\subsection{Task Coverage is Fragmented Across Countries}

Geographic disparities extend beyond dataset availability to the range of
NLP tasks represented. We measure each country's \textit{task breadth}, the
number of task categories represented, and \textit{top-3 task share}, the
proportion of represented dataset records concentrated in its three most
common tasks. We restrict the analysis to countries with at least 10
represented dataset records and classify portfolios with a top-3 share above 0.60 as \textit{specialist}, and the remainder as \textit{generalist}.

\begin{table}[t]
\centering
\scalebox{0.7}{
\begin{tabular}{lrrrl}
\toprule
\textbf{Country} &
\begin{tabular}[c]{@{}c@{}}\textbf{Represented}\\ \textbf{records}\end{tabular} &
\begin{tabular}[c]{@{}c@{}}\textbf{Task}\\ \textbf{breadth}\end{tabular} &
\begin{tabular}[c]{@{}c@{}}\textbf{Top-3}\\ \textbf{share}\end{tabular} &
\textbf{Portfolio} \\
\midrule
United States & 631 & 27 & 0.50 & Generalist \\
China         & 368 & 27 & 0.48 & Generalist \\
India         & 233 & 22 & 0.46 & Generalist \\
France        & 144 & 22 & 0.44 & Generalist \\
Egypt         & 58  & 15 & 0.53 & Generalist \\
\midrule
Mexico        & 49  & 14 & 0.61 & Specialist \\
Belgium       & 26  & 7  & 0.81 & Specialist \\
Bahrain       & 15  & 7  & 0.73 & Specialist \\
Israel        & 20  & 8  & 0.70 & Specialist \\
Oman          & 20  & 9  & 0.70 & Specialist \\
\bottomrule
\end{tabular}}
\caption{Examples of country-level task portfolios under explicit attribution.
Task breadth is the number of represented task categories, and top-3 share is
the proportion of a country's represented  dataset records concentrated in the three most common tasks. We label portfolios with top-3 share above 0.60 as specialist and the remainder as generalist. Countries with fewer than 10 represented dataset records are excluded.}
\label{tab:task_specialization}
\end{table}

Even among these better-represented countries, task coverage remains uneven.
Under explicit attribution, the median country spans 11 of 30 task categories,
while its three most common tasks account for 55.6\% of represented dataset records. Explicit+inferred attribution increases median breadth to 13 tasks, while concentration remains similar at 54.3\%.

Table~\ref{tab:task_specialization} illustrates this variation. Countries
such as the United States, China, India, and France have broad generalist
portfolios, whereas Belgium, Bahrain, Israel, and Oman are concentrated in
a much narrower set of tasks. Thus, geographic coverage does not imply broad task coverage: a country may be represented in NLP datasets but remain concentrated in only a small subset of tasks.

\subsection{Language Metadata Does Not Imply Geographic Coverage}

A central challenge in analyzing dataset geography is that represented-country
information is often unavailable or insufficiently documented. Datasets more
commonly report language, making language an appealing proxy for geographic
coverage even though the same language may span many countries and populations.

AtlasNLP addresses this gap by distinguishing explicit represented-country
attribution from additional inferred attribution. Inferred attribution was intentionally conservative, as language alone was not used to assign broadly distributed languages such as English, French, Spanish, Portuguese, Arabic, or Swahili to countries; such assignments required additional country-specific contextual evidence. Even under explicit+inferred attribution, however, a represented country could not be established for 9,956 of 13,462 Core dataset records (74.0\%). This substantial attribution gap illustrates why language metadata alone cannot resolve the geography of much of the NLP dataset ecosystem.

\begin{figure}[h]
  \includegraphics[width=\columnwidth]{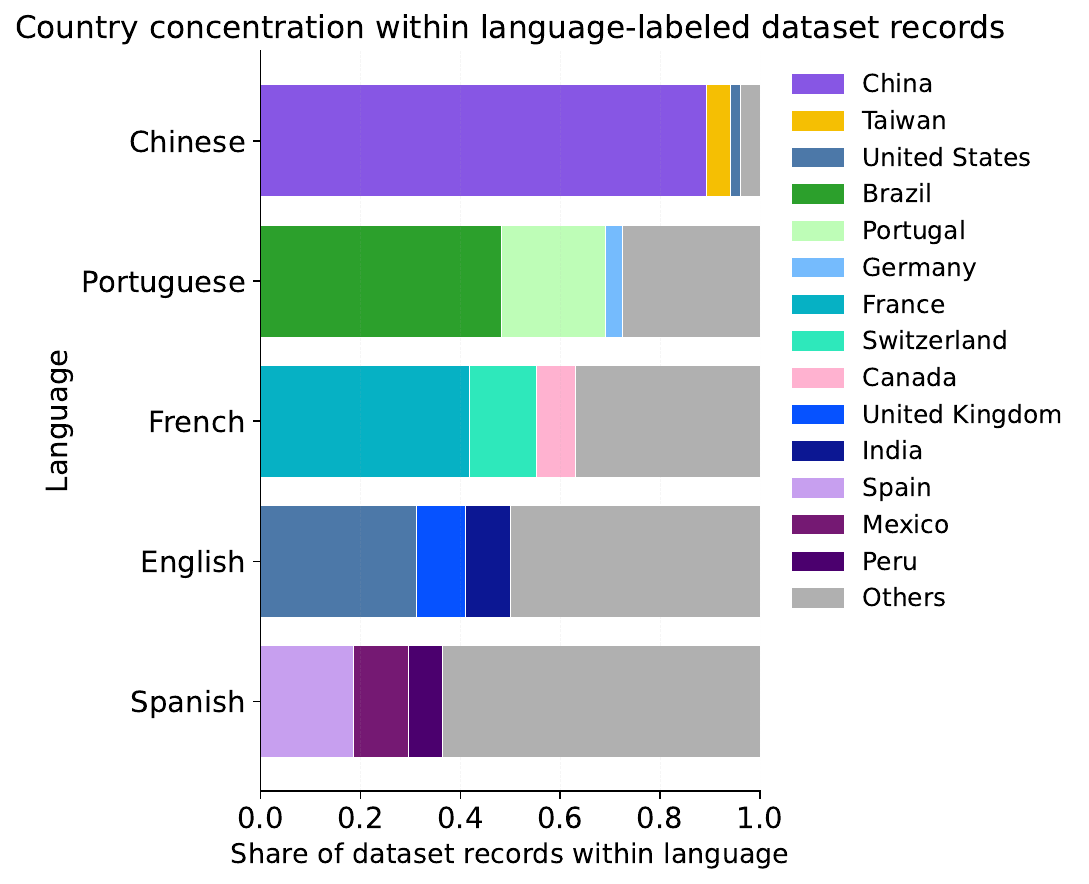}
  \caption{Country concentration within language-labeled dataset records under explicit attribution. Country representation varies substantially within widely used languages: some are dominated by one or a few countries, while others are more geographically distributed. Language coverage therefore does not directly indicate which countries are represented.}
  \label{fig:language_country_concentration}
\end{figure}

Figure~\ref{fig:language_country_concentration} shows substantial variation
in country concentration across widely used languages. Chinese-language
records are overwhelmingly associated with China (89.2\%). For Portuguese,
Brazil and Portugal together account for about 69\% of represented dataset records,
while France, Switzerland, and Canada account for about 63\% of French
records. English is more geographically distributed, yet the United States,
United Kingdom, and India still account for about half of represented dataset records.

These patterns show why language alone is an incomplete measure of geographic
representation. A language may span many countries without those countries
being represented equally, or at all, within NLP datasets. Explicit country-aware metadata is therefore necessary for observing population-level gaps that language metadata alone can obscure.

\section{Lessons Learned and Implications}

\paragraph{Document represented populations explicitly.}
A major challenge in constructing AtlasNLP was the absence of recoverable
country-level provenance. Even with explicit+inferred attribution, a
represented country could not be established for 74\% of AtlasNLP-Core dataset 
records. This does not imply that these datasets lack geographic context,
but that it is often not documented in a form that can be reliably recovered.
Dataset documentation should therefore include represented countries or
populations, geographic scope, and data provenance to support more transparent
and accurate coverage claims \cite{longpre2024data}.

\paragraph{Distinguish explicit from inferred geography.}
Plausible geographic signals should not be treated as equivalent to documented
provenance. Separating explicit and inferred attribution makes uncertainty
visible and allows geographic analyses to be evaluated under both strict and
broader assumptions.

\paragraph{Treat language as necessary but insufficient.}
Language remains essential for organizing NLP resources, but it should not be
treated as a proxy for geographic representation \cite{tonneau2024languages}.
Countries sharing a language can have very different levels of representation,
and language-labeled resources may remain concentrated in only a subset of the
populations that use that language.

\paragraph{Separate representation from production.}
Dataset geography is shaped by both who is represented and who produces the data. Distinguishing \textit{represented geography} from \textit{producer geography} makes institutional asymmetries in dataset creation more visible.

\paragraph{Track task coverage, not only dataset counts.}
Countries with some dataset coverage may still lack coverage across large parts of the NLP task space. Reporting task-level geographic coverage can help identify where dataset development is broad, narrow, or missing.

\paragraph{Make geographic metadata first-class dataset infrastructure.}
Large dataset platforms such as Hugging Face have improved resource visibility,
but population-level coverage remains difficult to audit without standardized
metadata for represented populations, geographic provenance, and producer
geography. AtlasNLP offers one schema for making these distinctions explicit
and geographic gaps easier to identify.

\section{Conclusion}

We introduced AtlasNLP, a country-aware atlas of NLP dataset contributions
that supports analysis of geographic representation, dataset production,
and task coverage. By distinguishing the populations represented in datasets
from the institutional locations where datasets are produced, and explicit
from inferred geographic attribution, AtlasNLP makes visible patterns that
are difficult to observe in language-centered resource collections.

Our analysis shows that NLP dataset coverage is highly uneven across countries
and tasks, production is geographically concentrated and asymmetric, and
language coverage does not imply geographic coverage. These findings highlight
limitations in current dataset documentation practices and show why geographic
diversity cannot be assessed reliably from language metadata or producer
geography alone. We release AtlasNLP at
\href{https://lit.eecs.umich.edu/AtlasNLP/index.html}{the AtlasNLP project site} and hope it supports future work on dataset transparency, geographic representation, and population-aware evaluation, while encouraging geographic metadata to become a first-class component of dataset documentation.


\section*{Limitations}

AtlasNLP is subject to limitations in metadata quality, coverage, and scope.
First, represented-country attribution remains incomplete and non-random.
Our precision-first strategy leaves dataset records unattributed when country
provenance cannot be established confidently, particularly for globally
distributed languages. Consequently, low or zero country coverage should be
interpreted as limited \textit{documented} representation within the ACL
literature, rather than evidence that no relevant NLP resources exist.

Second, AtlasNLP-Core consists of paper-level dataset-contribution records rather than fully deduplicated dataset entities. Related dataset records may describe extensions, compilations, or versions of the same underlying resource, for which entity-level deduplication is not always straightforward. AtlasNLP therefore measures documented dataset activity, not the number of unique datasets, dataset size, quality, or downstream evaluation capacity. Multi-country  dataset records may also contribute to several country totals. These counts should therefore be interpreted as indicators of research visibility and coverage rather than data volume.

Third, most Core annotations are automatically extracted. Validation against
AtlasNLP-Gold and targeted audits indicate high precision for the geographic
layers used in our analyses, but errors may remain, particularly for complex
multi-country scope. Gold also shares the AtlasNLP annotation schema and is
not a fully independent ground truth.

Fourth, the ModernBERT NLI filter used to identify candidate dataset papers
was not evaluated for population-wide recall across the full ACL Anthology.
AtlasNLP should therefore not be interpreted as an exhaustive census of all
dataset contributions in ACL, and we avoid temporal claims that would depend
on uniform recall across publication periods.

Fifth, producer geography is derived from author affiliations and represents
institutional location rather than researcher identity, nationality, or intent.
This distinction is especially important for diaspora researchers and
cross-border collaborations.

Finally, AtlasNLP-Core is intentionally scoped to ACL Anthology publications.
Datasets released only through Hugging Face, GitHub, industry repositories,
government portals, community archives, or non-ACL venues are therefore
underrepresented. Our normalized task schema also trades granularity for
cross-country comparability and may obscure locally specific or niche task
distinctions.

\section*{Ethical Considerations}

AtlasNLP makes country-level gaps in NLP dataset representation more visible,
but country-level analysis can also be overinterpreted. Countries are not
equivalent to cultures, languages, ethnic groups, or lived experiences, and
substantial variation exists within national boundaries. We therefore caution
against using AtlasNLP to rank countries, essentialize populations, or treat
country labels as complete representations of identity or culture.

Geographic attribution also carries uncertainty. Inferred country assignments
capture plausible geographic relevance under deliberately conservative rules;
they should not be interpreted as definitive claims about the identity of
dataset contributors or populations. For this reason, AtlasNLP preserves the
distinction between explicit and inferred attribution and uses explicit
provenance for its primary geographic analyses.

There is also a risk that evidence of underrepresentation could be used to
normalize poorer model performance for some regions rather than motivate its
improvement. Our goal is the opposite: to identify gaps in representation,
production, and task coverage so that future data collection and evaluation
can be more inclusive and accountable. Likewise, absence or low coverage in
AtlasNLP should not be interpreted as evidence that a country or community
lacks NLP resources.

Producer-country metadata should be interpreted carefully. It reflects
institutional affiliation, not researcher nationality, identity, or intent.
Externally produced datasets may result from collaboration, diaspora
scholarship, or resource sharing; our aim is not to privilege domestic
production, but to make production and representation patterns visible.

AtlasNLP-Gold was constructed through a collaborative research effort rather than paid crowdsourcing. Contributors compiled and validated entries using shared guidelines and were credited as coauthors when they completed the agreed curation and validation workload. Machine-assisted extensions followed the same human curation goals and guidelines. Gold was deliberately curated to broaden geographic coverage and should therefore not be interpreted as a natural sample of the global NLP dataset ecosystem.

\section*{Acknowledgments}
We thank the anonymous reviewers for their feedback, and the members of the Language and Information Technologies lab at the University of Michigan for the insightful discussions during the early stage of the project. This project was partially funded by 
a grant from OpenAI and 
a grant from the Survival and Flourishing Fund and a grant from the University of Michigan under the Strategic Initiative Fund. 
Any opinions, findings, and conclusions or recommendations expressed in this material are those of the authors and do not necessarily reflect the views of Open AI or the Survival and Flourishing Fund or the University of Michigan.

\bibliography{custom}

\appendix

\label{sec:appendix}

\section{Pipeline and Construction Details}
\label{app:pipeline}

\begin{figure*}[!t]
\centering

\begin{tikzpicture}[
    every node/.style={
        font=\scriptsize,
        align=center
    },
    gold/.style={
        draw,
        rounded corners,
        fill=orange!12,
        text width=4.8cm,
        minimum height=0.72cm,
        inner sep=4pt
    },
    core/.style={
        draw,
        rounded corners,
        fill=blue!8,
        text width=4.8cm,
        minimum height=0.72cm,
        inner sep=4pt
    },
    finalgold/.style={
        draw,
        rounded corners,
        fill=orange!20,
        very thick,
        text width=4.8cm,
        minimum height=0.82cm,
        inner sep=4pt
    },
    finalcore/.style={
        draw,
        rounded corners,
        fill=blue!18,
        very thick,
        text width=4.8cm,
        minimum height=0.82cm,
        inner sep=4pt
    },
    validation/.style={
        draw,
        rounded corners,
        fill=green!12,
        text width=11.4cm,
        minimum height=1.05cm,
        inner sep=5pt
    },
    analysis/.style={
        draw,
        rounded corners,
        fill=gray!8,
        text width=3.25cm,
        minimum height=1.08cm,
        inner sep=3pt,
        font=\scriptsize
    },
    arrow/.style={
        -{Latex[length=2.7mm,width=1.9mm]},
        line width=0.9pt
    }
]


\coordinate (G) at (-3.15,0);
\coordinate (C) at ( 3.15,0);


\node[
    font=\bfseries\normalsize,
    text=orange!80!black
] at ($(G)+(0,0)$)
{AtlasNLP-Gold};

\node[
    font=\bfseries\normalsize,
    text=blue!70!black
] at ($(C)+(0,0)$)
{AtlasNLP-Core};


\node[gold] (g1) at ($(G)+(0,-0.9)$)
{Contributor country assignment\\
197-entity ontology};

\node[gold] (g2) at ($(G)+(0,-2.05)$)
{Dataset search and annotation\\
ACL Anthology, Hugging Face, GitHub,\\
regional sources, etc.};

\node[gold] (g3) at ($(G)+(0,-3.25)$)
{Cross-validation\\
country, task, links, metadata, and provenance};

\node[gold] (g4) at ($(G)+(0,-4.40)$)
{Machine-assisted extension\\
and post-curation audit};

\node[finalgold] (g5) at ($(G)+(0,-5.65)$)
{\textbf{AtlasNLP-Gold}\\
1,480 human-curated entries\\
989 normalized dataset-name groups};


\node[core] (c1) at ($(C)+(0,-0.9)$)
{ACL Anthology bibliography\\
\textbf{119,963 records}};

\node[core] (c2) at ($(C)+(0,-2.05)$)
{Records with non-empty abstracts\\
\textbf{71,847}};

\node[core] (c3) at ($(C)+(0,-3.25)$)
{ModernBERT NLI filter\\
\textbf{20,277 candidates}\\
entailment score $>0.5$};

\node[core] (c4) at ($(C)+(0,-4.40)$)
{Role triage + GPT-4o-mini extraction\\
\textbf{18,035 successful initial extractions}};

\node[core] (c5) at ($(C)+(0,-5.65)$)
{400-record human audit\\
dataset role + represented-country attribution};

\node[core] (c6) at ($(C)+(0,-6.90)$)
{Audit-driven full re-evaluation\\
stricter dataset-role and\\
geographic-attribution criteria};

\node[finalcore] (c7) at ($(C)+(0,-8.15)$)
{\textbf{AtlasNLP-Core}\\
\textbf{13,462 primary dataset-contribution records}};


\node[validation] (v1) at (0,-9.75)
{\textbf{Validation and final quality checks}\\[1pt]
Gold $\leftrightarrow$ Core:
225 high-confidence dataset matches
\quad $\cdot$ \quad
400-record post-revision audit verification\\
100-record producer-country audit:
precision 96.9\%, recall 97.7\%};


\node[analysis] (a1) at (-4.25,-11.45)
{\textbf{Explicit geography}\\
2,447 dataset records\\
4,421 country-record associations\\
158 countries};

\node[analysis] (a2) at (0,-11.45)
{\textbf{Explicit + inferred}\\
3,506 dataset records\\
6,001 country-record associations\\
168 countries};

\node[analysis] (a3) at (4.25,-11.45)
{\textbf{Producer geography}\\
13,224 dataset records\\
98.2\% coverage};


\draw[arrow] (g1.south) -- (g2.north);
\draw[arrow] (g2.south) -- (g3.north);
\draw[arrow] (g3.south) -- (g4.north);
\draw[arrow] (g4.south) -- (g5.north);


\draw[arrow] (c1.south) -- (c2.north);
\draw[arrow] (c2.south) -- (c3.north);
\draw[arrow] (c3.south) -- (c4.north);
\draw[arrow] (c4.south) -- (c5.north);
\draw[arrow] (c5.south) -- (c6.north);
\draw[arrow] (c6.south) -- (c7.north);


\draw[arrow]
    (g5.south) -- ([xshift=-3.15cm]v1.north);

\draw[arrow]
    (c7.south) -- ([xshift=3.15cm]v1.north);


\draw[arrow]
    ([xshift=-4.25cm]v1.south) -- (a1.north);

\draw[arrow]
    (v1.south) -- (a2.north);

\draw[arrow]
    ([xshift=4.25cm]v1.south) -- (a3.north);

\end{tikzpicture}

\caption{
Overview of AtlasNLP construction, audit, and validation.
AtlasNLP-Core proceeds from ACL candidate selection through initial automated
extraction and audit-driven re-evaluation, yielding 13,462 final primary
dataset-contribution records. AtlasNLP-Gold contains 1,480 human-curated
entries corresponding to 989 normalized dataset-name groups. Final geographic
layers distinguish explicit attribution from the broader explicit+inferred
sensitivity analysis.
}
\label{fig:pipeline}
\end{figure*}
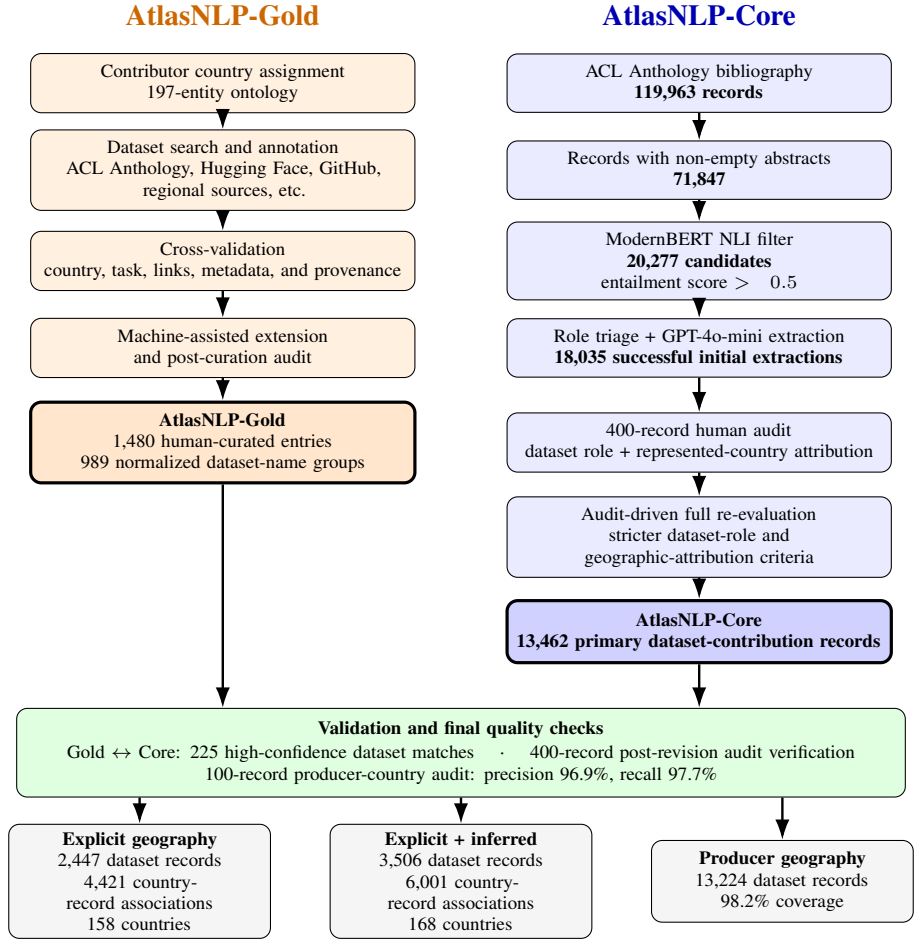

This appendix provides additional details on the construction, audit, and
validation of AtlasNLP-Core and AtlasNLP-Gold. Figure~\ref{fig:pipeline}
summarizes the full pipeline. AtlasNLP-Core is constructed in two stages:
an initial large-scale extraction from ACL Anthology papers, followed by
audit-driven refinement of dataset eligibility and geographic attribution.
AtlasNLP-Gold is constructed through human curation, cross-validation, and
subsequent audit and refinement. The sections below describe the metadata
schema, task taxonomy, country-attribution policy, language normalization,
automated extraction, post-extraction audit, producer-country extraction,
and construction of the country-level analysis layers.

The initial ACL pipeline begins with 119,963 ACL Anthology records, of which
71,847 contain non-empty abstracts. ModernBERT-base-NLI assigns 23,674 papers
an entailment label for the hypothesis that the paper proposes a dataset;
20,277 also exceed the 0.5 entailment threshold and enter the extraction
pipeline. Role triage and LLM-assisted metadata extraction produce 18,035
successful initial paper-level extractions. A subsequent 400-record human
audit identifies recurring dataset-role and geographic-attribution errors
and motivates a full re-evaluation of these records. The final AtlasNLP-Core
contains 13,462 primary dataset-contribution records.

\subsection{Metadata Schema}
\label{app:schema}

AtlasNLP uses a shared conceptual schema for human-curated and automatically
extracted dataset records. The schema captures dataset identity and provenance,
dataset role, task and language coverage, represented and producer geography,
and additional dataset characteristics. The final Core release also retains
audit and normalization fields used to document eligibility and geographic
attribution. Table~\ref{tab:app_schema} summarizes the principal metadata
fields.

\begin{table*}[!t]
\centering
\footnotesize
\setlength{\tabcolsep}{6pt}
\renewcommand{\arraystretch}{1.08}

\begin{tabular}{p{0.19\textwidth}p{0.74\textwidth}}
\toprule
\textbf{Field} & \textbf{Description} \\
\midrule

Dataset name &
Name of the dataset, benchmark, resource, or dataset compilation. \\

Paper link &
Paper URL or citation associated with the dataset contribution. \\

Dataset link &
Official dataset, benchmark, repository, GitHub, or Hugging Face URL. \\

Dataset relationship / role &
Relationship between the paper and target dataset: introduced, materially
extended, newly compiled, reused from prior work, not identifiable, or
ambiguous. This field determines AtlasNLP-Core eligibility. \\

Task category &
Normalized NLP task category from the AtlasNLP taxonomy. \\

Language(s) &
Dataset languages or language varieties, normalized and audited when possible. \\

Represented country &
Country or population represented by the dataset content, participants,
sources, or collection context. \\

Country attribution &
Geographic evidence layer: \textit{explicit}, \textit{inferred}, or
unattributed when evidence is insufficient. \\

Producer country &
Institutional location of dataset creators, derived from author affiliations
for ACL-derived records. \\

Modality &
Dataset modality, including text-only and multimodal resources. \\

Multiple-choice format &
Whether the dataset or benchmark uses a multiple-choice format. \\

Synthetic &
Whether content is machine-translated, model-generated, or otherwise synthetic,
when recoverable. \\

License &
Dataset license, when available. \\

Provenance / evidence &
Source information supporting extracted metadata, including page- and
quote-level evidence where available. \\

Audit / normalization fields &
Eligibility decisions, attribution status, normalized country lists, processing
status, and related post-audit quality-control fields. \\

\bottomrule
\end{tabular}

\caption{
Principal metadata fields used in AtlasNLP. The final Core release also
retains audit and normalization fields documenting dataset eligibility and
geographic-attribution decisions.
}
\label{tab:app_schema}
\end{table*}

The schema separates \textit{represented country}, which captures the
geographic population or context represented by a dataset, from
\textit{producer country}, which captures where dataset production is
institutionally located. It also separates dataset identity from
\textit{dataset relationship}: related papers may introduce, materially extend,
compile, or reuse the same underlying resource. Final AtlasNLP-Core eligibility
is therefore defined at the level of the paper--dataset contribution rather
than by dataset-name deduplication alone.

\subsection{Task Taxonomy}
\label{app:task_taxonomy_section}

AtlasNLP uses a normalized task taxonomy derived from recent ACL and EMNLP
thematic areas. AtlasNLP-Core contains 30 task categories, while
AtlasNLP-Gold additionally includes \textit{Raw Corpus}, yielding 31
categories across AtlasNLP overall. Table~\ref{tab:app_task_taxonomy}
lists the final taxonomy and short labels used in figures.

\begin{table*}[t]
\centering
\small
\resizebox{0.85\textwidth}{!}{%
\begin{tabular}{ll}
\toprule
\textbf{Task category} & \textbf{Short label} \\
\midrule
Code Models & Code Models \\
Computational Social Science, Cultural Analytics, NLP for Social Good & CSS / NLP4SG \\
Dialogue and Interactive Systems & Dialogue \\
Discourse, Pragmatics, and Reasoning & Discourse / Reasoning \\
Efficient / Low-Resource Methods for NLP, LLM Efficiency & Low-resource / Efficiency \\
Ethics, Bias, and Fairness & Ethics / Bias \\
Generalizability and Transfer & Transfer \\
Hierarchical Structure Prediction & Hierarchical \\
Human-AI Interaction / Cooperation, Human-Centered NLP & Human-Centered NLP \\
Information Extraction, Retrieval, and Text Mining & IE / Retrieval \\
Interpretability, Model Analysis, Transparency, and Explainability & Interpretability \\
Language Modeling & Language Modeling \\
Linguistic Theories, Cognitive Modeling, and Psycholinguistics & Psycholinguistics \\
Machine Learning for NLP & ML for NLP \\
Machine Translation & MT \\
Mathematical, Symbolic, and Logical Reasoning in NLP & Math / Logic \\
Multilinguality and Language Diversity & Multilinguality \\
Multimodality and Language Grounding to Vision, Robotics, and Beyond & Multimodality \\
Natural Language Generation & NLG \\
Neurosymbolic Approaches to NLP & Neurosymbolic \\
NLP Applications & NLP Applications \\
Phonology, Morphology, and Word Segmentation & Phonology / Morphology \\
Question Answering & QA \\
Raw Corpus & Raw Corpus \\
Retrieval-Augmented Language Models & RAG \\
Safety and Alignment in LLMs & Safety \\
Semantics: Lexical and Sentence Level, Textual Inference & Semantics / Inference \\
Sentiment Analysis, Stylistic Analysis, and Argument Mining & Sentiment / Style / Argument \\
Speech Recognition, Text-to-Speech, and Spoken Language Understanding & Speech \\
Summarization & Summarization \\
Syntax: Tagging, Chunking, and Parsing & Syntax \\
\bottomrule
\end{tabular}}
\caption{AtlasNLP-Core+Gold task taxonomy. AtlasNLP-Core contains 30 ACL/EMNLP-derived categories; AtlasNLP-Gold adds the human-curated \textit{Raw Corpus} category. Short labels are used in compact visualizations.}
\label{tab:app_task_taxonomy}
\end{table*}

\subsection{Country Attribution Rules}
\label{app:country_attribution}

AtlasNLP uses a standardized set of 197 geopolitical entities, consisting of
the 193 UN member states and two UN observer states, together with Taiwan and
Kosovo. Represented-country attribution is separated from producer geography
and is classified into three confidence levels: \textit{explicit},
\textit{inferred}, and \textit{unattributed}.

\begin{itemize}

    \item \textbf{Explicit attribution.}
    A country is assigned explicitly when dataset documentation directly
    connects the dataset content, participants, data sources, collection
    process, target population, or benchmark scope to that country. Examples
    include data collected in a named country, a benchmark explicitly designed
    for a country's population, or source material drawn from a clearly
    identified national context.

    \item \textbf{Inferred attribution.}
    A country is assigned as inferred when geographic relevance is plausible
    from indirect but geographically informative evidence, such as a
    country-specific language variety, a locally defined community, or a data
    source with a sufficiently narrow geographic scope. Inferred attribution
    is retained separately from explicit attribution and is used only in the
    broader sensitivity analyses.

    \item \textbf{Unattributed.}
    A dataset record remains unattributed when available evidence is
    insufficient to support either explicit or conservative inferred
    geography. Broadly distributed languages, author affiliations, incidental
    country mentions, and the geographic location of the producing institution
    are not by themselves treated as evidence of represented country.

\end{itemize}

This distinction is important because \textit{represented country} and
\textit{producer country} capture different concepts. Author affiliation may
establish producer geography but does not establish which population the
dataset represents. The post-extraction audit identified affiliation-based
assignment, language-only assignment, and incidental country mentions as
recurring sources of erroneous represented-country attribution; the final
pipeline therefore applies these exclusions explicitly.

For multi-country datasets, all countries supported by the available evidence
are retained. A single dataset record can therefore contribute to multiple
country-record associations, while remaining one dataset record in analyses
that do not expand by represented country.

Synthetic datasets follow the same attribution policy. Producer country is
derived from the institution responsible for dataset creation, while
represented country is assigned only when the dataset documentation identifies
a target population or geographic context. For example, a model-generated
Swahili dataset explicitly designed to simulate Kenyan speakers can be
attributed to Kenya, whereas synthetic language-model output with no stated
target population remains geographically unattributed.

\subsection{Language Normalization and Inference}
\label{app:language_inference}

Language metadata is often noisy, inconsistent, or incomplete. AtlasNLP
therefore normalizes language metadata before using it for analysis or
geographic inference. The normalization process includes whitespace and
formatting cleanup, alias resolution, ISO-code expansion, preservation of
locale-specific labels, separation of canonical languages from language
varieties and sign languages, and removal of invalid or non-language entries.
Programming-language labels and other artifacts are excluded from the audited
language inventory.

Language normalization is conceptually separate from country attribution.
A normalized language label does not automatically imply a represented
country. Instead, language may contribute to the \textit{inferred} geographic
layer only when it provides a sufficiently specific geographic signal.
Explicit country attribution is never assigned from language alone.

Table~\ref{tab:app_language_policy} summarizes the final conservative
language-to-country policy.

\begin{table}[!t]
\centering
\footnotesize
\setlength{\tabcolsep}{4pt}
\renewcommand{\arraystretch}{1.08}

\begin{tabular}{p{0.31\columnwidth}p{0.59\columnwidth}}
\toprule
\textbf{Language signal} & \textbf{Treatment} \\
\midrule

Country-specific variety or locale &
May support \textit{inferred} attribution when the language label itself
provides a sufficiently narrow geographic signal, such as Egyptian Arabic
for Egypt. \\

Narrow geographic association &
May support conservative inferred attribution when the language or speech
community is strongly associated with a limited geographic context. If
multiple countries remain plausible, AtlasNLP does not force assignment to a
single country. \\

Language plus contextual evidence &
May support inferred attribution when language is accompanied by additional
country-specific evidence, such as a locally defined community, source, or
collection context. \\

Broadly distributed language &
No country is assigned from language alone. This includes languages such as
English, French, Spanish, Portuguese, Arabic, and Swahili; country-specific
context is required. \\

Script-only or ambiguous label &
No represented country is assigned when the available metadata identifies
only a script, unresolved variant, or otherwise insufficient geographic
signal. \\

\bottomrule
\end{tabular}

\caption{
Conservative treatment of language signals in represented-country inference.
Language may support the inferred layer when geographic evidence is
sufficiently specific, but language alone never produces explicit country
attribution.
}
\label{tab:app_language_policy}
\end{table}

The final attribution layers preserve this uncertainty explicitly. Among the
13,462 AtlasNLP-Core dataset records, 2,447 have explicit represented-country
attribution, 1,059 have inferred-only attribution, and 9,956 remain
unattributed. Thus, the broader explicit+inferred sensitivity layer contains
3,506 dataset records. Inferred attribution is not synonymous with
language-based mapping: it may also reflect other indirect but geographically
informative evidence described in Appendix~\ref{app:country_attribution}.

The audited-language outputs are retained independently of these geographic
decisions. The language audit was performed on the extracted records and then
restricted to the final eligible Core and reviewed Gold collections, preserving
the original language-level audit decisions after the post-extraction dataset
audit. Final language inventory statistics are reported in
Appendix~\ref{app:language_inventory}.

\subsection{Automated Extraction for AtlasNLP-Core}
\label{app:automated_extraction}

AtlasNLP-Core is constructed from ACL Anthology publications through a
multi-stage candidate-selection and metadata-extraction pipeline. We begin
with 119,963 ACL Anthology bibliography records spanning 1952--2025.
Of these, 71,847 contain non-empty abstracts and are eligible for the
dataset-paper filtering stage.

We use ModernBERT-base-NLI to evaluate whether each paper's title and abstract
support the hypothesis:

\begin{quote}
``This paper proposes a dataset.''
\end{quote}

The NLI model assigns 23,674 papers an entailment label. We retain the
20,277 papers whose entailment score also exceeds 0.5 as candidate dataset
papers for downstream extraction.

Candidate papers are then processed using a staged extraction pipeline.
For each paper, the pipeline retrieves the paper PDF, extracts full text
page-by-page, performs an initial rule-based dataset-role triage, and applies
schema-constrained GPT-4o-mini extraction to recover dataset metadata.

The extraction stage records dataset identity, dataset relationship, task
category, language, licensing, represented-country signals, modality,
synthetic status, format, and related metadata. Automatically extracted
fields are accompanied by page- and quote-level evidence when available.
Records requiring additional clarification are routed to a targeted second
LLM pass using relevant paper excerpts rather than being accepted directly.

The initial role-triage step excludes 905 candidate papers before LLM extraction. A further 1,337 candidates
fail during PDF retrieval, parsing, schema validation, or related extraction
steps. The remaining 18,035 papers yield successful initial paper-level
extractions:

\[
20{,}277
=
905
+
1{,}337
+
18{,}035.
\]

These 18,035 records form the \textit{initial extraction set}, rather than
the final AtlasNLP-Core collection. As described in
Appendix~\ref{app:audit}, a subsequent 400-record human audit identified
remaining dataset-role and represented-country attribution errors. The resulting audit criteria were then applied across the full initial extraction set using the separate post-audit re-evaluation pipeline described in Appendix~\ref{app:audit}, yielding the final 13,462 AtlasNLP-Core primary dataset-contribution records.

\subsection{Metadata Extraction Prompt}
\label{app:metadata_prompt}

The LLM-assisted extraction stage uses a schema-constrained prompt designed
to minimize unsupported inference and preserve auditable provenance. The model
is instructed to return valid JSON, mark unavailable fields as
\texttt{Not stated}, and provide page-level evidence for extracted metadata.
This prompt governs the initial extraction stage; dataset-role eligibility and
represented-country attribution are subsequently re-evaluated using the
post-extraction audit procedure described in Appendix~\ref{app:audit}.

A simplified version of the extraction prompt is shown below.

\begin{lstlisting}[
    basicstyle=\ttfamily\scriptsize,
    breaklines=true,
    columns=fullflexible
]
System prompt:
You extract dataset metadata from the provided paper text
conservatively and auditably.

Rules:
- Do not guess. Do not infer beyond the available evidence.
- If a field is not supported by the paper text, output
  "Not stated".
- Every non-"Not stated" field MUST have an evidence item
  with a page number and a short quote (<= 20 words).
- Output MUST be valid JSON only.
- If multiple datasets are described, identify the primary
  target dataset when possible; otherwise flag ambiguity.
- All required fields must be present.

User prompt:
Return JSON with this structure:

{
  paper:{...},
  dataset:{...},
  evidence:[...],
  quality:{
    flags:[],
    needs_stage2:boolean
  },
  run:{...}
}

Allowed task categories:
[ALLOWED_TASK_CATEGORIES]

Paper ID:
[PAPER_ID]

Paper URL:
[PAPER_URL]

Prompt version:
[PROMPT_VERSION]

Paper text with page markers:
[PAPER_TEXT_WITH_PAGE_MARKERS]
\end{lstlisting}

\subsection{Post-Extraction Audit and Refinement}
\label{app:audit}

The 18,035 successful initial extractions were designed for high recall and
still contained cases in which the paper did not make a primary dataset
contribution or in which represented-country evidence was insufficiently
grounded in dataset provenance. We therefore conducted a human audit of 400
randomly sampled records from the initial extraction set.

The audit reviewed two issues separately. First, reviewers determined the
relationship between the paper and the target dataset: whether the paper
introduced a new dataset, materially extended an existing dataset, constructed
a new dataset compilation, reused an existing resource, or did not support a
confident target-dataset identification. Second, reviewers evaluated
represented-country attribution against evidence about the dataset itself,
rather than author affiliation, language alone, or incidental country
mentions.

The audit identified recurring failure modes in both dimensions. In
particular, some initial records described reused datasets rather than primary
dataset contributions, while some represented-country assignments reflected
the location of authors or institutions rather than the population represented
by the dataset. Other errors arose from broadly distributed languages,
incidental country mentions, incomplete multi-country scope, and ambiguous
dataset identity. These findings were used to tighten both dataset-role and
geographic-attribution criteria before re-evaluating the full 18,035-record
initial extraction set.

\paragraph{Full re-evaluation implementation.}
We used the pinned gpt-5.4-mini-2026-03-17 model for both stages of the audit-driven re-evaluation. The first stage used low reasoning effort over the full paper text to classify the paper–dataset relationship and extract per-country provenance evidence. Records with candidate country evidence were passed to a separate verification stage using medium reasoning effort, which classified each proposed country as explicit, inferred, or unsupported. The verifier did not receive the original AtlasNLP country or producer-country assignments. Final country layers were constructed deterministically from the verified evidence.

\paragraph{Final dataset eligibility.}
A record is retained in AtlasNLP-Core only when the target dataset can be
matched with sufficient confidence and the paper makes one of three primary
dataset contributions: (i) introducing the dataset, (ii) materially extending
an existing dataset, or (iii) constructing a new dataset compilation. Mere
reuse of an existing dataset does not qualify.

Table~\ref{tab:app_final_eligibility} summarizes the resulting final Core
composition.

\begin{table}[!t]
\centering
\footnotesize
\setlength{\tabcolsep}{5pt}
\renewcommand{\arraystretch}{1.08}

\begin{tabular}{lr}
\toprule
\textbf{Final eligible relationship} & \textbf{Dataset records} \\
\midrule
Dataset introduced in paper & 12,205 \\
Existing dataset materially extended & 819 \\
New dataset compilation constructed & 438 \\
\midrule
\textbf{Final AtlasNLP-Core} & \textbf{13,462} \\
\bottomrule
\end{tabular}

\caption{
Composition of final AtlasNLP-Core by paper--dataset relationship.
Records are retained only when the target dataset is confidently matched and
the paper makes a primary dataset contribution.
}
\label{tab:app_final_eligibility}
\end{table}

Of the 4,573 initial extractions not retained in final Core, 4,071 involve
reuse of an existing dataset, 408 do not yield a sufficiently identifiable
target dataset, and 42 involve unresolved ambiguity among multiple datasets.
An additional 48 records have a potentially eligible dataset relationship but
lack a sufficiently confident match to the target dataset, and four records
could not be processed successfully during re-evaluation. Thus, the reduction
from 18,035 initial extractions to 13,462 final Core dataset records reflects
dataset-role and identity refinement rather than removal based on geographic
coverage.

\paragraph{Geographic refinement.}
Dataset eligibility is determined independently of whether represented-country
metadata can be recovered. Eligible records remain in AtlasNLP-Core even when
no country can be established. For geographic analyses, represented-country
evidence is instead assigned to the separate \textit{explicit},
\textit{inferred}, or \textit{unattributed} layers described in
Appendix~\ref{app:country_attribution}. This separation prevents uncertainty in
geographic provenance from being conflated with whether a paper contributes a
dataset.

The same 400-record audit is subsequently used to evaluate whether the revised
pipeline addresses the failure modes observed during review. Post-revision
audit results, including country-attribution corrections and abstention
behavior, are reported in Appendix~\ref{app:validation}.

\subsection{Producer-Country Extraction}
\label{app:producer_extraction}

AtlasNLP-Core also records producer geography, defined as the institutional
location associated with dataset production. Producer country is derived from
author affiliations rather than researcher nationality, identity, or
represented-country information. It is therefore treated as conceptually
separate from represented-country attribution.

The producer-side pipeline combines rule-based affiliation extraction with
targeted LLM fallback. For each ACL paper, the pipeline retrieves affiliation
information from the paper, including first-page text where available, and
uses deterministic methods to identify institution names, explicit country
mentions, email domains, and other affiliation cues. Country assignment
follows a hierarchy of evidence that includes direct country mentions,
institution--country mappings, and domain-based signals. Extracted country
names are normalized to the common 197-entity AtlasNLP ontology.

When rule-based extraction leaves producer metadata missing or ambiguous, a
targeted LLM fallback is applied. Trigger conditions include missing country
or institution information and unresolved or conflicting affiliation
evidence. Rule-based results are retained when sufficiently supported, while
the fallback is used to resolve or fill remaining gaps.

In the final AtlasNLP-Core collection, producer-country metadata is available
for 13,224 of 13,462 dataset records (98.2\%). We independently evaluate this
pipeline on 100 human-labeled records. At the country-association level,
producer-country extraction achieves 96.9\% precision and 97.7\% recall, and
the primary producer country is recovered in all 100 audited records. Full
audit results are reported in Appendix~\ref{app:validation}.

\subsection{Country--Task Matrix Construction}
\label{app:matrix_construction}

Country-task matrices are constructed from the final 13,462-record
AtlasNLP-Core collection after country and task normalization. Each Core
dataset record has one normalized task category. For geographic analyses,
records are expanded over all represented countries supported by the relevant
attribution layer, so that a multi-country dataset contributes once to each
supported country while remaining a single dataset record in the underlying
Core collection.

We construct two versions of the matrix. The primary matrix uses only
\textit{explicit} represented-country attribution. This layer contains 2,447
dataset records, which expand to 4,421 country-record associations covering
158 countries. The sensitivity matrix additionally includes
\textit{inferred} geography, yielding 3,506 dataset records and 6,001
country-record associations across 168 countries.

For comparability across geographic layers, both matrices are represented over
the complete AtlasNLP-Core analysis space of 197 geopolitical entities and 30
task categories, including zero-count rows and columns where applicable. Each
matrix cell therefore records the number of country-record associations for a
given country and normalized task.

The explicit matrix contains 1,231 nonzero country-task cells out of
5,910 possible cells, leaving 79.2\% of the matrix empty. Under the broader
explicit+inferred attribution layer, 1,480 cells are nonzero and 75.0\%
remain empty. Thus, the broader attribution layer increases observed
geographic coverage but does not substantially alter the overall sparsity of
country-task representation.

These matrices underpin the country-task coverage and task-portfolio analyses
reported in the main paper. Unless otherwise stated, primary results use the
explicit matrix, while the explicit+inferred matrix is used as a sensitivity
analysis.

\section{Data Transparency}
\label{app:data_transparency}

This appendix summarizes the controlled geographic and language vocabularies
used in AtlasNLP, coverage under the different geographic-attribution layers,
and the audited language inventory.

\subsection{Country Ontology and Coverage}
\label{app:country_ontology}

AtlasNLP uses a standardized ontology of 197 geopolitical entities:
193 UN member states, the two UN non-member observer states, and Taiwan and
Kosovo. The observer states are represented in the ontology as Palestine and
Vatican City. Country names are normalized to this ontology before analysis,
reducing variation from aliases, abbreviations, alternate spellings, and
sub-national references.

\begin{table}[!h]
\centering
\footnotesize
\setlength{\tabcolsep}{4pt}
\renewcommand{\arraystretch}{1.08}

\begin{tabular}{@{}lcc@{}}
\toprule
\textbf{Subset} &
\textbf{Explicit} &
\textbf{Explicit + inferred} \\
\midrule

AtlasNLP-Core &
158 / 197 &
168 / 197 \\

AtlasNLP-Gold &
192 / 197 &
197 / 197 \\

\bottomrule
\end{tabular}

\caption{
Geopolitical-entity coverage under the two represented-country attribution
layers. Primary analyses use explicit attribution; explicit+inferred coverage
is reported as a sensitivity analysis.
}
\label{tab:app_country_coverage}
\end{table}

Table~\ref{tab:app_country_ontology} lists the complete ontology.

\begin{table*}[th]
\centering
\small
\resizebox{\textwidth}{!}{%
\begin{tabular}{llll}
\toprule
\textbf{Country/entity} & \textbf{Country/entity} & \textbf{Country/entity} & \textbf{Country/entity} \\
\midrule
Afghanistan & Ecuador & Luxembourg & São Tomé and Príncipe \\
Albania & Egypt & Madagascar & Saudi Arabia \\
Algeria & El Salvador & Malawi & Senegal \\
Andorra & Equatorial Guinea & Malaysia & Serbia \\
Angola & Eritrea & Maldives & Seychelles \\
Antigua and Barbuda & Estonia & Mali & Sierra Leone \\
Argentina & Eswatini & Malta & Singapore \\
Armenia & Ethiopia & Marshall Islands & Slovakia \\
Australia & Fiji & Mauritania & Slovenia \\
Austria & Finland & Mauritius & Solomon Islands \\
Azerbaijan & France & Mexico & Somalia \\
Bahamas & Gabon & Micronesia & South Africa \\
Bahrain & Gambia & Moldova & South Korea \\
Bangladesh & Georgia & Monaco & South Sudan \\
Barbados & Germany & Mongolia & Spain \\
Belarus & Ghana & Montenegro & Sri Lanka \\
Belgium & Greece & Morocco & Sudan \\
Belize & Grenada & Mozambique & Suriname \\
Benin & Guatemala & Myanmar & Sweden \\
Bhutan & Guinea & Namibia & Switzerland \\
Bolivia & Guinea-Bissau & Nauru & Syria \\
Bosnia and Herzegovina & Guyana & Nepal & Taiwan \\
Botswana & Haiti & Netherlands & Tajikistan \\
Brazil & Honduras & New Zealand & Tanzania \\
Brunei & Hungary & Nicaragua & Thailand \\
Bulgaria & Iceland & Niger & Timor-Leste \\
Burkina Faso & India & Nigeria & Togo \\
Burundi & Indonesia & North Korea & Tonga \\
Cabo Verde & Iran & North Macedonia & Trinidad and Tobago \\
Cambodia & Iraq & Norway & Tunisia \\
Cameroon & Ireland & Oman & Turkey \\
Canada & Israel & Pakistan & Turkmenistan \\
Central African Republic & Italy & Palau & Tuvalu \\
Chad & Jamaica & Palestine & Uganda \\
Chile & Japan & Panama & Ukraine \\
China & Jordan & Papua New Guinea & United Arab Emirates \\
Colombia & Kazakhstan & Paraguay & United Kingdom \\
Comoros & Kenya & Peru & United States \\
Congo (Congo-Brazzaville) & Kiribati & Philippines & Uruguay \\
Costa Rica & Kosovo & Poland & Uzbekistan \\
Cote d'Ivoire & Kuwait & Portugal & Vanuatu \\
Croatia & Kyrgyzstan & Qatar & Vatican City \\
Cuba & Laos & Romania & Venezuela \\
Cyprus & Latvia & Russia & Vietnam \\
Czechia & Lebanon & Rwanda & Yemen \\
Democratic Republic of the Congo & Lesotho & Saint Kitts and Nevis & Zambia \\
Denmark & Liberia & Saint Lucia & Zimbabwe \\
Djibouti & Libya & Saint Vincent and the Grenadines &  \\
Dominica & Liechtenstein & Samoa &  \\
Dominican Republic & Lithuania & San Marino &  \\
\bottomrule
\end{tabular}}
\caption{
AtlasNLP country ontology used for normalization and coverage analysis.
The ontology contains 193 UN member states, the two UN non-member observer
states, Taiwan, and Kosovo.}
\label{tab:app_country_ontology}
\end{table*}

Coverage depends on the geographic-attribution layer being considered.
Table~\ref{tab:app_country_coverage} therefore reports coverage separately
for explicit attribution and for the broader explicit+inferred sensitivity
layer. AtlasNLP-Core provides explicit represented-country evidence for
158 of the 197 entities; including conservative inferred attribution increases
coverage to 168. AtlasNLP-Gold provides substantially broader geographic
coverage, with explicit evidence for 192 entities and explicit+inferred
coverage for all 197.

The difference between ontology membership and observed coverage is
intentional. Countries without attributed dataset records remain part of the
197-entity analysis space and therefore appear as zero-count cases rather than
being dropped from country-level analyses.

\subsection{Language Inventory and Audit}
\label{app:language_inventory}

AtlasNLP records both raw language metadata and audited language labels.
Raw fields may contain ISO codes, aliases, dialect or locale labels, script
names, formatting variants, and non-language artifacts. We therefore
normalize and audit language metadata before using it for language-level
analysis. The normalization and conservative use of language in geographic
inference are described in Appendix~\ref{app:language_inference}.

The language audit was conducted on the source collections and then aligned
to the final post-audit Core and Gold records. All final records were
successfully matched to their audited-language annotations. Programming-language
artifacts identified during the audit, including \textit{Ada} and
\textit{ABAP}, are excluded from the published language inventory.

\begin{center}
\footnotesize
\setlength{\tabcolsep}{8pt}
\renewcommand{\arraystretch}{1.08}
\begin{tabular}{@{}lr@{}}
\toprule
\textbf{Inventory} & \textbf{Audited languages} \\
\midrule
AtlasNLP-Core & 1,239 \\
AtlasNLP-Gold & 269 \\
AtlasNLP overall & 1,245 \\
\bottomrule
\end{tabular}
\end{center}

The Core and Gold inventories overlap substantially but are not identical.
Six audited languages occur in Gold but not in final Core:
\textit{Basaa}, \textit{Guinea-Bissau Creole}, \textit{Hiri Motu},
\textit{Inuinnaqtun}, \textit{Myene}, and \textit{Nzebi}. The union of the
two final inventories therefore contains 1,245 audited languages.

These language counts describe normalized language coverage rather than
geographic coverage. In particular, the presence of a language label does not
imply that a represented country can be established for the corresponding
dataset record.

\subsection{Language-to-Country Mapping Transparency}
\label{app:language_mapping_transparency}

Language metadata is not treated as direct evidence of represented-country
coverage. In particular, AtlasNLP does not convert normalized language labels
into countries through a general language--country lookup. Many languages span
multiple countries, and assigning every dataset in a language to all countries
where that language is spoken would substantially overstate geographic
representation.

Language can contribute only to the \textit{inferred} geographic layer when
the available label or accompanying context provides a sufficiently specific
geographic signal. For example, a country-specific variety such as Egyptian
Arabic may support inferred attribution to Egypt. By contrast, broadly
distributed languages such as English, French, Spanish, Portuguese, Arabic,
or Swahili do not support country assignment on their own. Script-only labels,
unresolved varieties, and cases compatible with several countries likewise
remain geographically unattributed unless additional evidence is available.

Language-derived evidence never produces \textit{explicit} attribution.
Explicit represented-country assignments require direct provenance connecting
the dataset content, participants, sources, collection process, or target
population to a country. The complete inference policy is described in
Appendix~\ref{app:language_inference}. This separation allows AtlasNLP to
retain rich language metadata without treating language coverage as equivalent
to geographic coverage.

\section{Validation and Quality Checks}
\label{app:validation}

We evaluate AtlasNLP-Core through three complementary validation procedures.
First, we compare final Core metadata against independently curated
AtlasNLP-Gold records that can be aligned with high confidence. Second, we
revisit the 400-record human audit used to motivate the post-extraction
refinement and measure whether the revised pipeline corrects the observed
dataset-role and represented-country attribution errors. Third, we conduct a
separate 100-record human audit of producer-country extraction. Together,
these checks evaluate dataset identity, task assignment, represented-country
attribution, abstention behavior, and producer geography.

\subsection{Gold-Core Alignment}
\label{app:gold_core_validation}

AtlasNLP-Gold and AtlasNLP-Core are constructed through different pipelines,
and a shared paper does not necessarily imply a shared dataset record. Papers
may introduce or discuss multiple datasets, and dataset names can vary across
sources. We therefore perform validation only after establishing a
high-confidence dataset-level alignment rather than treating paper-level
overlap alone as a match.

We begin with the 1,480 post-audit Gold entries marked as complete. Among
these, 339 unique papers have ACL Anthology links and can therefore be
compared directly with the ACL-derived Core pipeline. Of these papers, 299
are recovered in the initial 18,035-record extraction set, and 267 remain
after the final dataset-role and identity refinement that produces the
13,462-record Core. The 32 recovered papers that do not remain in final Core
consist of 28 cases classified as reuse of an existing dataset and four for
which the target dataset could not be identified with sufficient confidence.

\begin{center}
\footnotesize
\setlength{\tabcolsep}{6pt}
\renewcommand{\arraystretch}{1.08}
\begin{tabular}{@{}lr@{}}
\toprule
\textbf{Alignment stage} & \textbf{Count} \\
\midrule
ACL-linked Gold papers & 339 \\
Recovered in initial extraction set & 299 \\
Retained in final Core & 267 \\
High-confidence dataset matches & 225 \\
\bottomrule
\end{tabular}
\end{center}

Within the 267 papers retained in final Core, we then establish
dataset-level correspondence using conservative identity criteria. We first
match the normalized Gold and Core dataset names within the same ACL paper.
This yields 203 high-confidence matches. When normalized names do not match,
we accept a record only if Gold and Core provide the same non-empty dataset
URL within the same paper, yielding 22 additional matches. We do not use
paper-URL agreement alone to infer dataset identity.

The resulting validation set therefore contains 225 high-confidence
Gold-Core dataset matches: 203 based on exact normalized dataset-name
agreement and 22 based on exact dataset-URL agreement. These matched records
form the reference population for the represented-country and task comparisons
reported in Appendix~\ref{app:gold_core_metadata_validation}.

\subsection{Represented-Country and Task Validation}
\label{app:gold_core_metadata_validation}

We evaluate represented-country agreement on the 225 high-confidence
Gold--Core dataset matches described above. Because the final pipeline
distinguishes \textit{explicit} from \textit{inferred} geography, we perform
the comparison separately under the two attribution layers used in the main
analysis.

For each layer, we distinguish between (i) the number of Gold reference
records with geographic evidence and (ii) the subset for which Core also
produces a represented-country attribution. Agreement metrics are computed
conditional on Core making a geographic prediction. A Core abstention is
therefore treated as missing geographic coverage rather than as an incorrect
country assignment.

\begin{center}
\footnotesize
\setlength{\tabcolsep}{5pt}
\renewcommand{\arraystretch}{1.08}

\resizebox{0.88\columnwidth}{!}{%
\begin{tabular}{@{}lcc@{}}
\toprule
\textbf{Metric} &
\textbf{Explicit} &
\textbf{Explicit + inferred} \\
\midrule

Gold reference records &
181 &
225 \\

Core predicts geography &
87 (48.1\%) &
138 (61.3\%) \\

Any country overlap &
97.7\% &
97.8\% \\

Exact country-set match &
66.7\% &
68.8\% \\

Country-assignment precision &
94.7\% &
94.0\% \\

\bottomrule
\end{tabular}%
}
\end{center}

Under the explicit-only comparison, 181 matched Gold records have explicit
represented-country evidence. Core assigns explicit geography to 87 of these
records. Conditional on an assignment, the Core and Gold country sets overlap
in 97.7\% of cases, and 66.7\% have an exact country-set match. At the
individual country-assignment level, precision is 94.7\%.

Under the broader explicit+inferred comparison, all 225 matched Gold records
have geographic evidence in the corresponding Gold layer. Core assigns
explicit or inferred geography to 138 records. Conditional on an assignment,
97.8\% have at least one country in common with Gold and 68.8\% have an exact
country-set match. Country-assignment precision is 94.0\%.

Exact-set agreement is deliberately strict. For a multi-country dataset, the
metric is counted as correct only when the complete predicted country set
matches the Gold country set exactly; both omitted and additional countries
cause the record to fail exact-set agreement. The substantially higher
overlap scores therefore indicate that most non-exact cases reflect differences
in multi-country scope rather than completely unrelated geographic
attribution.

\paragraph{Task agreement.}
Task labels are compared only when the Gold record has a single task label that
maps directly to the final Core taxonomy. Of the 225 high-confidence dataset
matches, 219 satisfy this criterion. Core assigns the same normalized task
category as Gold in 175 cases, yielding 79.9\% exact task agreement.

These results support the use of the automated metadata for aggregate
country- and task-level analysis while also motivating the conservative
treatment of missing geography and the explicit separation of inferred
attribution from the primary explicit layer.

\subsection{Post-Revision Audit Verification}
\label{app:audit_validation}

We next return to the 400 records used in the post-extraction human audit and
evaluate their outcomes under the revised pipeline. The purpose of this check
is different from the Gold--Core comparison above: rather than measuring
agreement with an independent dataset collection, it tests whether the
specific dataset-role and geographic-attribution failure modes identified
during the audit persist after full re-evaluation.

The human audit classifies 287 of the 400 sampled records as eligible primary
dataset contributions, 112 as ineligible, and one as a processing failure.

\begin{center}
\footnotesize
\setlength{\tabcolsep}{8pt}
\renewcommand{\arraystretch}{1.08}
\begin{tabular}{@{}lr@{}}
\toprule
\textbf{Human-audit outcome} & \textbf{Records} \\
\midrule
Eligible primary contribution & 287 \\
Excluded & 112 \\
Processing failure & 1 \\
\midrule
Total & 400 \\
\bottomrule
\end{tabular}
\end{center}

\paragraph{Country-attribution corrections.}
The original audit labels 187 represented-country assignments as incorrect.
After the audit-driven re-evaluation, 177 of these cases (94.7\%) no longer
remain in the explicit geographic layer. Similarly, among 85 cases for which
the human audit judged the geographic connection to be inferred rather than
explicit, 83 (97.6\%) are not incorrectly promoted to explicit attribution.

A particularly important test concerns abstention. Among the 287
dataset-eligible audit records, 98 cases have no represented country supported
by the human review. The revised pipeline leaves 97 of these 98 records
geographically unattributed. This indicates that the revised procedure largely
distinguishes unsupported geography from extraction failure rather than forcing
a country assignment whenever country-related information appears in a paper.

\paragraph{Observed failure modes.}
The audit revealed several recurring reasons for unsupported country
assignments. For example, the initial extraction assigned the United States to
the \textit{Non-Cooperative Guess What?! Dataset}, reflecting the creators'
University of Pittsburgh affiliation; however, the paper identifies the
participants only as native English speakers and does not establish their
country. The revised record therefore remains geographically unattributed.
Similarly, \textit{MADial-Bench} was initially associated with China, the
United Kingdom, and Australia, but the dataset is synthetically constructed
without direct evidence that these countries constitute represented
populations; the revised pipeline abstains from country attribution.
\textit{EventRelBench} provides another example in which an incidental
country signal led to an initial Japan assignment despite insufficient
dataset-level provenance; this assignment is likewise removed.

These examples illustrate why country-attribution errors cannot be corrected
simply by improving country-name extraction. The relevant distinction is
whether a geographic signal describes the dataset's represented population or
content rather than an author affiliation, conference location, incidental
mention, language association, or other contextual information.

The post-revision audit therefore supports the conservative design used in
the final analyses: unsupported assignments are preferentially removed or
downgraded from explicit attribution, while uncertainty is preserved through
the separate inferred and unattributed layers.

\subsection{Producer-Country Validation}
\label{app:producer_validation}

We independently evaluate producer-country extraction on a manually reviewed
sample of 100 AtlasNLP-Core records. Human reviewers inspect author
affiliations and record the producer-country set associated with each paper.
The automated output is then compared against this reference at both the
country-association and record levels.

Across the 100 audited records, the extractor predicts 130 producer-country
associations, of which 126 are correct. The human reference contains 129
associations, of which 126 are recovered. This yields 96.9\% precision and
97.7\% recall.

At the record level, the complete producer-country set matches the human
reference exactly for 93 of 100 papers. Importantly, the primary producer
country is recovered in all 100 audited records, including cases where
additional secondary affiliations cause the full country set to differ.

\begin{center}
\footnotesize
\setlength{\tabcolsep}{7pt}
\renewcommand{\arraystretch}{1.08}

\resizebox{0.82\columnwidth}{!}{%
\begin{tabular}{@{}lr@{}}
\toprule
\textbf{Producer-country audit metric} & \textbf{Result} \\
\midrule
Country-assignment precision & 96.9\% \\
Country-assignment recall & 97.7\% \\
Exact country-set match & 93 / 100 \\
Primary producer country recovered & 100 / 100 \\
\bottomrule
\end{tabular}%
}
\end{center}

The audit also allows us to inspect the two extraction paths separately.
Among the 60 records resolved primarily through deterministic affiliation
parsing, 51 have an exact producer-country-set match. All 40 records routed
through the targeted LLM fallback have an exact country-set match in the audit.
These results suggest that the fallback is useful for affiliation structures
that remain unresolved after rule-based parsing, while the combined pipeline
provides high producer-country coverage without relying exclusively on LLM
inference.

Because producer geography is derived from author affiliations, this audit
evaluates only the producer side of AtlasNLP. It does not validate
represented-country attribution, which is evaluated separately in
Appendices~\ref{app:gold_core_metadata_validation} and
\ref{app:audit_validation}.

\section{Human Annotation and Validation Process}
\label{app:gold}
\label{app:human_annotation}

This appendix describes the curation and validation process used to construct
AtlasNLP-Gold. Gold was designed to broaden geographic coverage beyond the
ACL-derived Core collection, establish the annotation schema, and provide an
independently curated reference for validation. The collection originated
through collaborative human curation and was subsequently extended and audited
with machine assistance under the same curation criteria.

\subsection{Contributor Assignment}

Collaborators selected countries from a shared assignment sheet. Countries
were grouped into high-, medium-, and low-resource levels based on expected
NLP dataset availability. The resource levels were used to balance annotation
workload rather than to characterize countries intrinsically.

\begin{center}
\footnotesize
\renewcommand{\arraystretch}{1.10}

\resizebox{0.88\columnwidth}{!}{%
\begin{tabular}{@{}lp{0.56\columnwidth}@{}}
\toprule
\textbf{Resource level (credit)} & \textbf{Expected workload} \\
\midrule

High (5) &
Many datasets; higher deduplication burden \\

Medium (3) &
Moderate expected dataset availability \\

Low (1) &
Few or potentially no identifiable datasets \\

\bottomrule
\end{tabular}%
}
\end{center}

Collaborators were asked to select enough countries to reach a target workload
and were encouraged to choose countries across multiple regions and resource
levels. When possible, contributors selected countries for which they had
geographic, linguistic, or domain familiarity. This design aimed to distribute
the curation workload while supporting careful country-specific search.

\subsection{Dataset Search Procedure}

For each assigned country, collaborators searched for existing NLP datasets
using academic and public dataset sources. Suggested sources included ACL
Anthology, Hugging Face, Papers With Code, GitHub, Kaggle, Google Scholar,
Semantic Scholar, national data portals, institutional repositories, community
projects, and regional dataset collections. Contributors prioritized official
dataset sources such as repository pages, benchmark websites, Hugging Face
pages, GitHub repositories, DOIs, and associated papers.

A shared task--country matrix was used to reduce duplicate work. Before
entering a dataset, contributors checked whether the resource had already been
recorded. Existing entries were cross-checked rather than entered independently
as new resources; newly identified datasets were added to the contributor's
entry sheet and shared tracking matrix.

For lower-resource settings, contributors could also report cases in which no
relevant dataset was identified after search. These reports documented the
sources searched and supported the broader goal of distinguishing an
unsuccessful search from an unexamined country. They are not interpreted as
evidence that no NLP resource exists for that country.

\subsection{Human Data Entry}

Human-curated entries followed a shared AtlasNLP schema. Contributors recorded
dataset name, task category, language, represented country, dataset and paper
links, modality, multiple-choice format, synthetic status, license, provenance
notes, and geographic-attribution evidence.

Gold was initially organized around dataset--country annotation instances.
Multi-country resources could therefore be associated with more than one
country while retaining a common dataset identity. For final reporting, dataset
names are additionally normalized into dataset-name groups; these normalized
groups are used only as an approximate identity summary rather than as a claim
of complete entity-level deduplication.

\subsection{Task Annotation}

Contributors assigned datasets to task categories using a shared guide based
on ACL and EMNLP thematic areas. The guide included broad task categories and
example subtasks. For example, information extraction included named entity
recognition, relation extraction, event extraction, keyphrase extraction,
document retrieval, and knowledge-base population. Question answering included
extractive, abstractive, open-domain, multi-hop, multiple-choice, and
conversational QA. Multimodality included image captioning, visual question
answering, video--text alignment, and related language-grounded tasks.

If a dataset did not fit an existing category, contributors could propose a
new category. These additions were subsequently reviewed and normalized to the
final AtlasNLP taxonomy described in Appendix~\ref{app:task_taxonomy_section}.

\subsection{Human Country Attribution}

Represented-country annotation followed the same conceptual distinction used
in the final AtlasNLP analyses. Contributors prioritized direct evidence that
the dataset content, participants, data sources, collection setting, or target
population was associated with a country. Such evidence supports
\textit{explicit} attribution.

When direct country information was unavailable, geographically informative
but indirect evidence could support an \textit{inferred} attribution. Language
alone was treated conservatively. Broadly distributed languages such as
English, French, Spanish, Portuguese, Arabic, and Swahili were not considered
sufficient for country attribution without additional country-specific
evidence. Cases lacking adequate explicit or conservative inferred evidence
remained unattributed.

For multi-country datasets, annotators retained all countries supported by the
available evidence rather than forcing a single primary geographic label.
Country-attribution decisions were recorded separately from producer geography.

\subsection{Cross-Validation}

After initial compilation, entries were redistributed for secondary
validation. Validators performed targeted checks rather than re-annotating
every source from scratch. Validation focused on four areas:

\begin{itemize}

    \item \textbf{Dataset and link checks:}
    Validators confirmed that dataset names corresponded to the linked
    resources and attempted to repair broken or incorrect links when possible.

    \item \textbf{Core metadata checks:}
    Validators checked country, language, task, and geographic-attribution
    evidence for consistency with the cited resource.

    \item \textbf{Missing-value checks:}
    Missing fields were completed when the relevant information could be
    established with high confidence.

    \item \textbf{Formatting and normalization checks:}
    Validators corrected invalid controlled-vocabulary values, inconsistent
    country names, and formatting errors.

\end{itemize}

Validators recorded comments and changes in provenance or notes fields and
marked entries complete after validation.

\subsection{Country Name Normalization}

Country names were normalized to the 197-entity AtlasNLP ontology described in
Appendix~\ref{app:country_ontology}. Common corrections included mapping
``UAE'' to ``United Arab Emirates,'' ``America'' to ``United States,''
``Czech Republic'' to ``Czechia,'' ``Republic of Congo'' to
``Congo (Congo-Brazzaville),'' and ``Ivory Coast'' to
``Cote d'Ivoire.''

Where the ontology required country-level representation, sub-national labels
were normalized to the corresponding geopolitical entity; for example, Wales
and Scotland were mapped to the United Kingdom and Puerto Rico to the United
States. For multi-country entries, normalization was applied to individual
country labels while preserving the remaining country set.

\subsection{Machine-Assisted Extension and Final Audit}
\label{app:gold_extension}

Following the original collaborative curation, AtlasNLP-Gold was extended and
re-audited with machine assistance. These stages followed the same underlying
curation goals and geographic-attribution criteria as the original annotation
process. Machine assistance was used to surface candidate evidence, identify
records requiring reconsideration, and support consistency checks; it did not
change the definition of represented-country evidence.

The subsequent audit focused especially on records for which country
attribution depended on indirect evidence, dataset identity was unclear, or
the available provenance did not support the existing geographic assignment.
Unsupported assignments were removed, while geographically plausible but
indirectly supported cases were retained separately as inferred attribution.
This process also aligned Gold country names, attribution labels, and language
metadata with the final controlled vocabularies used in Core.

\subsection{AtlasNLP-Gold Output}

The final AtlasNLP-Gold collection contains \textbf{1,480 curated entries}
corresponding to \textbf{989 normalized dataset-name groups}. Gold provides
explicit represented-country coverage for 192 of the 197 AtlasNLP entities;
including conservative inferred attribution expands this to all 197 entities.

Because Gold was deliberately curated to broaden geographic coverage, its
country distribution should not be interpreted as a naturally occurring
sample of the NLP dataset ecosystem. Its primary roles are complementary
coverage, metadata transparency, and independent reference data for validating
AtlasNLP-Core.

\section{Dataset Examples}
\label{app:dataset_examples}

Table~\ref{tab:dataset_examples} provides representative examples of the
geographic metadata cases distinguished in AtlasNLP: explicit multi-country
coverage, separation of represented and producer geography, conservative
abstention when represented-country evidence is unavailable, and inferred
geography for a synthetic resource.

\begin{table*}[!t]
\centering
\scriptsize
\setlength{\tabcolsep}{3pt}
\renewcommand{\arraystretch}{1.10}

\resizebox{0.96\textwidth}{!}{%
\begin{tabular}{
    >{\bfseries}p{0.14\linewidth}
    p{0.20\linewidth}
    p{0.20\linewidth}
    p{0.20\linewidth}
    p{0.20\linewidth}
}
\toprule
\textbf{Metadata field} &
\textbf{SALT} &
\textbf{AfriSpeech-Dialog} &
\textbf{ToxASCII} &
\textbf{Code-170k-lingala} \\
\midrule

Subset &
AtlasNLP-Gold &
AtlasNLP-Core &
AtlasNLP-Core &
AtlasNLP-Gold \\

Task category &
Machine Translation &
Dialogue and Interactive Systems &
Ethics, Bias, and Fairness &
Code Models \\

Language(s) &
English; Swahili; Luganda; Runyankole; Acholi; Lugbara; Ateso &
English &
English &
Lingala \\

Represented country &
Uganda; Kenya; Tanzania &
Nigeria; Kenya; South Africa &
Unattributed &
Congo (Congo-Brazzaville); Democratic Republic of the Congo; Angola;
Central African Republic \\

Producer country &
Not consistently available in Gold &
United States &
France &
Not consistently available in Gold \\

Country attribution &
Explicit &
Explicit &
Unattributed &
Inferred \\

Synthetic &
No &
No &
No &
Yes \\

Illustrates &
Explicit multi-country regional resource &
Represented--producer geographic distinction &
Producer metadata without supported represented geography &
Synthetic resource with inferred geographic relevance \\

\bottomrule
\end{tabular}%
}

\caption{
Representative AtlasNLP records illustrating common geographic metadata
cases. Gold sources extend beyond ACL Anthology and therefore do not
consistently contain producer-country metadata. Inferred geography is kept
separate from the explicit attribution used in primary analyses.
}
\label{tab:dataset_examples}
\end{table*}

\section{Hugging Face Comparison}
\label{app:huggingface}

Hugging Face is the largest public repository for NLP datasets, with over
900,000 datasets, including approximately 450,000 associated with text
modalities. However, many datasets lack the structured metadata required for
systematic analysis. Language and task annotations are missing in over
80-85\% of records, and no fields capture represented country or represented
population. Although an \texttt{uploader\_region} field is widely populated,
its distribution suggests that it reflects account defaults rather than
dataset provenance. Datasets linked to research papers tend to be more
consistently documented, but their proportion has declined as platform growth
has shifted toward industry and individual contributors.

These patterns reflect a trade-off between scale and structure. ACL-derived
dataset records provide more consistent metadata but more limited coverage,
while Hugging Face offers much greater scale without standardized annotations
for geographic representation. This lack of structured metadata highlights
the need for resources such as AtlasNLP, which organize existing dataset
contributions using a standardized metadata framework and can inform the
design of future dataset platforms.

\section{Robustness and Sensitivity Analyses}
\label{app:sensitivity}

Primary geographic analyses in the paper use only explicitly supported
represented-country attribution. To assess whether the main conclusions depend
on this conservative choice, we repeat key analyses using the broader
explicit+inferred attribution layer. This section also reports additional
robustness checks addressing multi-country resources, dataset identity, and
pipeline recovery.

\subsection{Sensitivity to Geographic-Attribution Layer}
\label{app:attribution_sensitivity}

Including conservative inferred geography increases the number of Core dataset
records with represented-country metadata from 2,447 to 3,506 and expands
observed country coverage from 158 to 168 of the 197 AtlasNLP entities.
Despite this increase, the principal geographic patterns remain similar.

\begin{center}
\footnotesize
\setlength{\tabcolsep}{4pt}
\renewcommand{\arraystretch}{1.08}

\resizebox{0.94\columnwidth}{!}{%
\begin{tabular}{@{}lcc@{}}
\toprule
\textbf{Statistic} &
\textbf{Explicit} &
\textbf{Explicit + inferred} \\
\midrule

Dataset records with geography &
2,447 &
3,506 \\

Countries represented &
158 &
168 \\

Empty country--task cells &
79.2\% &
75.0\% \\

Infrastructure Pearson $r$ &
0.65 &
0.70 \\

Infrastructure Spearman $\rho$ &
0.61 &
0.67 \\

Countries in task-portfolio analysis &
77 &
89 \\

Median task breadth &
11 &
13 \\

Median top-3 task share &
55.6\% &
54.3\% \\

\bottomrule
\end{tabular}%
}
\end{center}

Country-task coverage remains highly sparse: 79.2\% of cells are empty under
explicit attribution and 75.0\% remain empty after inferred geography is
included. The association between represented-record availability and research
infrastructure is also similar and slightly stronger under the broader layer,
with Pearson correlation increasing from approximately 0.65 to 0.70 and
Spearman correlation from 0.61 to 0.67.

Task-portfolio patterns are likewise stable. Among countries with at least
10 represented dataset records, the number of eligible countries increases
from 77 to 89, while median task breadth increases from 11 to 13 of the
30 Core task categories. Median concentration in the three most common tasks
changes little, from 55.6\% to 54.3\%.

Production-representation asymmetry also persists. Without a minimum-count
restriction, 60 of 89 countries (67.4\%) under explicit attribution and
65 of 96 countries (67.7\%) under explicit+inferred attribution have content
self-representation below 0.5. The primary results use the more conservative
minimum-count analysis, under which 39 of 62 countries with at least 10 represented records with producer metadata and 10 producer–representation associations are primarily represented through records produced by institutions outside the country.

Altogether, these comparisons indicate that adding conservative inferred
geography increases coverage but does not materially change the main
conclusions regarding country-task sparsity, production-representation
asymmetry, research-infrastructure alignment, or task concentration.

\subsection{Additional Robustness Checks}
\label{app:additional_robustness}

\paragraph{Broad multi-country records.}
Because records spanning many countries can contribute disproportionately to expanded country counts, we repeat the represented-country analysis after excluding records attributed to more than 20 countries. Under explicit attribution, only 10 of 2,447 attributed records (0.4\%) exceed this threshold, accounting for 234 of 4,421 country-record associations (5.3\%). After removing them, 4,187 associations remain across 154 countries. The top-10 represented-country set is unchanged, and country rankings remain highly stable (Spearman $\rho=.991$). The top-10 share increases from 48.6\% to 50.4\%, indicating that very broad multi-country records slightly dilute, rather than produce, the observed concentration.

\paragraph{Exact-name sensitivity.}
AtlasNLP-Core uses paper-level dataset-contribution records rather than globally resolved dataset entities because aliases, versions, extensions, and generic dataset names make full entity resolution ambiguous. As a limited sensitivity check, we collapse records sharing the same exact normalized dataset-name key. Of 13,462 Core records, 13,451 have a usable normalized key, corresponding to 13,131 distinct name groups. Within the explicit geographic layer, this collapse reduces 4,421 country-record associations to 4,399 name-group--country associations across 2,427 geographically attributed name groups. The top-10 represented-country set remains identical; across the baseline top 50 countries, rankings correlate at $\rho=.999$, with a maximum rank shift of two positions. The top-10 share changes by only 0.05 percentage points (48.59\% to 48.53\%). Thus, repeated exact normalized-name matches do not materially drive the headline geographic concentration results. This check is intentionally narrower than full dataset-entity resolution.

\begin{figure}[h]
  \includegraphics[width=\columnwidth]
  {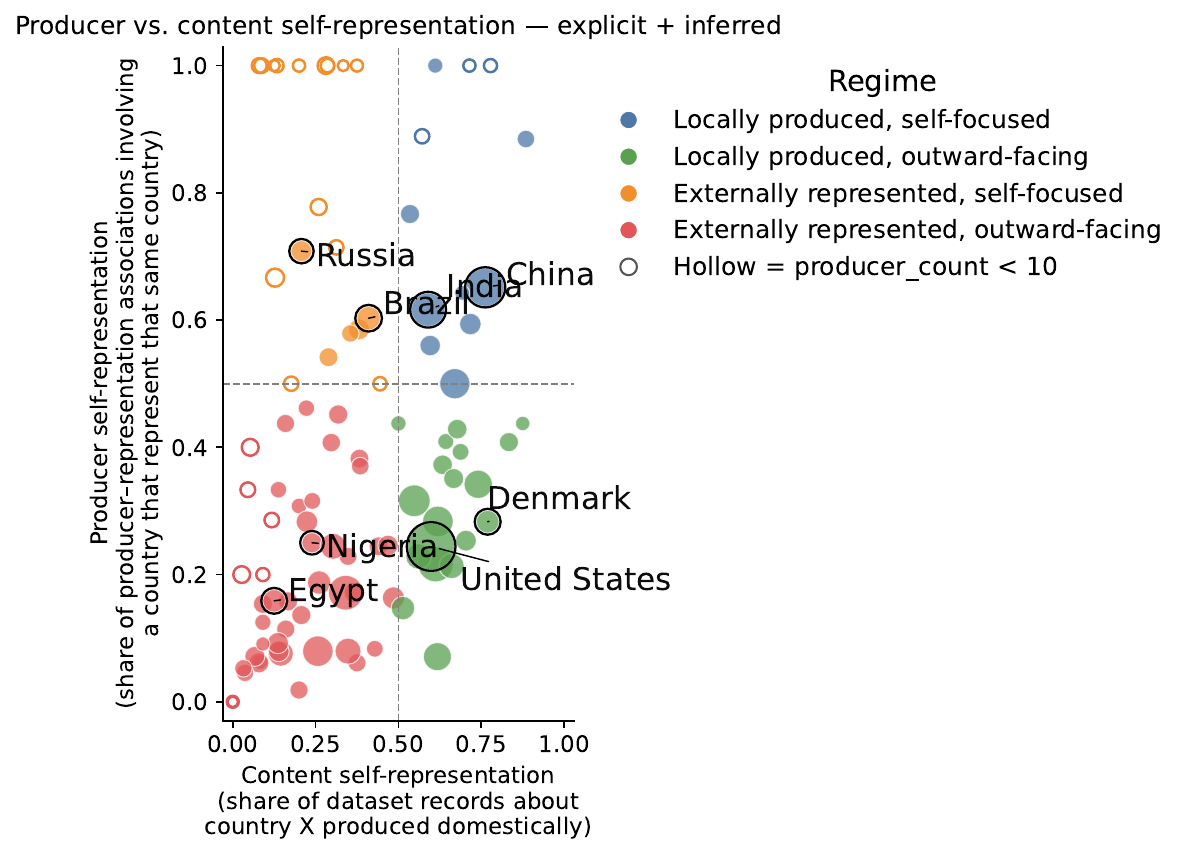}
  \caption{Dataset production and representation by country under explicit+inferred attribution. The x-axis measures the share of datasets about a country produced domestically; the y-axis measures the share of producer-representation associations involving a country that represent itself. Dashed lines define four production-representation regimes. Point size reflects dataset volume; hollow points indicate countries with fewer than 10 producer associations.
}
  \label{fig:app_represented_producer}
\end{figure}

\begin{figure}[h]
  \includegraphics[width=0.9\columnwidth]{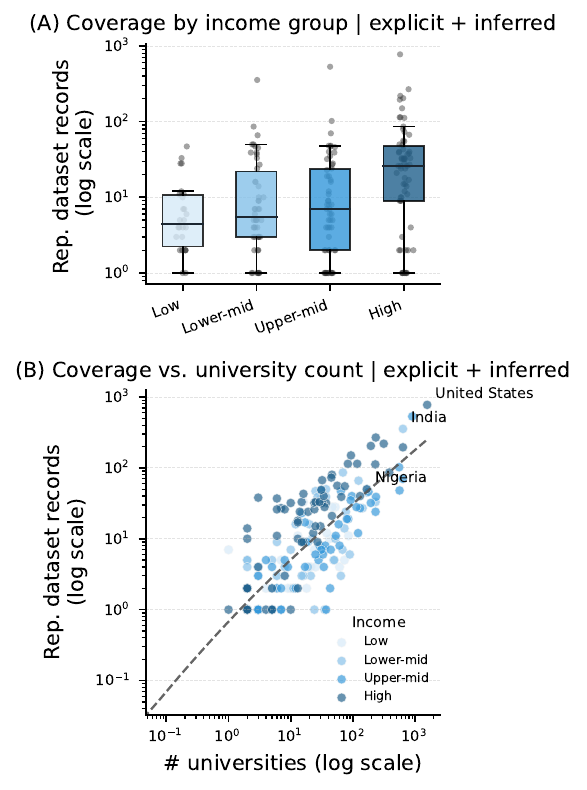}
\caption{
Relationship between represented-country dataset records and research
infrastructure under explicit+inferred attribution.
(A) Coverage by World Bank income group.
(B) Coverage increases with national university counts on log-transformed
values. Pearson $r\approx .70$ and Spearman $\rho\approx .67$.
Color indicates income group. Best viewed in color.
}
  \label{fig:app_infrastructure_alignment}
\end{figure}

\begin{figure}[h]
  \includegraphics[width=\columnwidth]{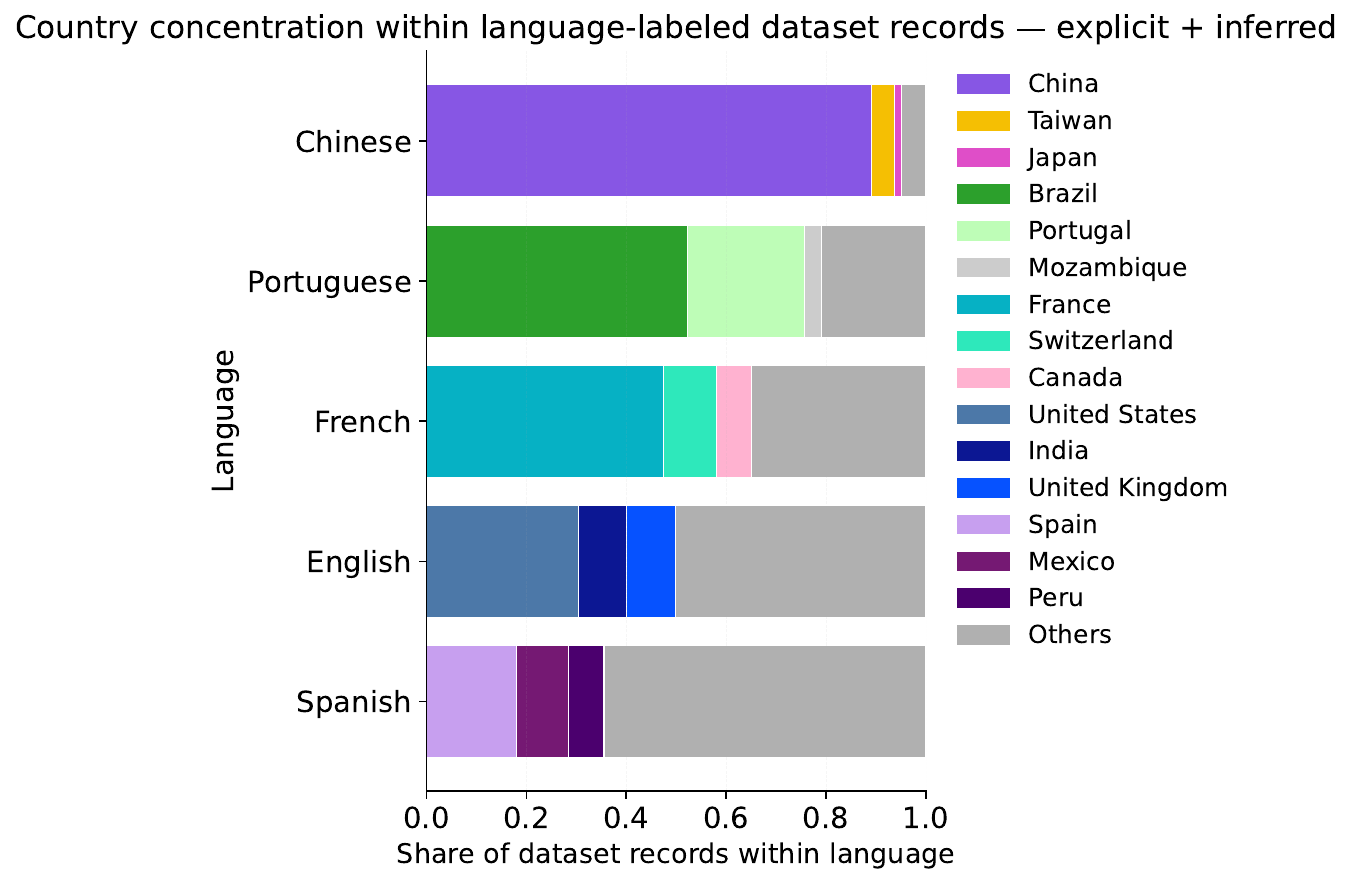}
  \caption{Country concentration within language-labeled dataset records under explicit+inferred attribution. Country representation varies substantially within widely used languages: some are dominated by one or a few countries, while others are more geographically distributed. Language coverage therefore does not directly indicate which countries are represented.}
  \label{fig:app_language_country_concentration}
\end{figure}

\begin{figure*}[h]
\centering
  \includegraphics[width=0.97\textwidth]{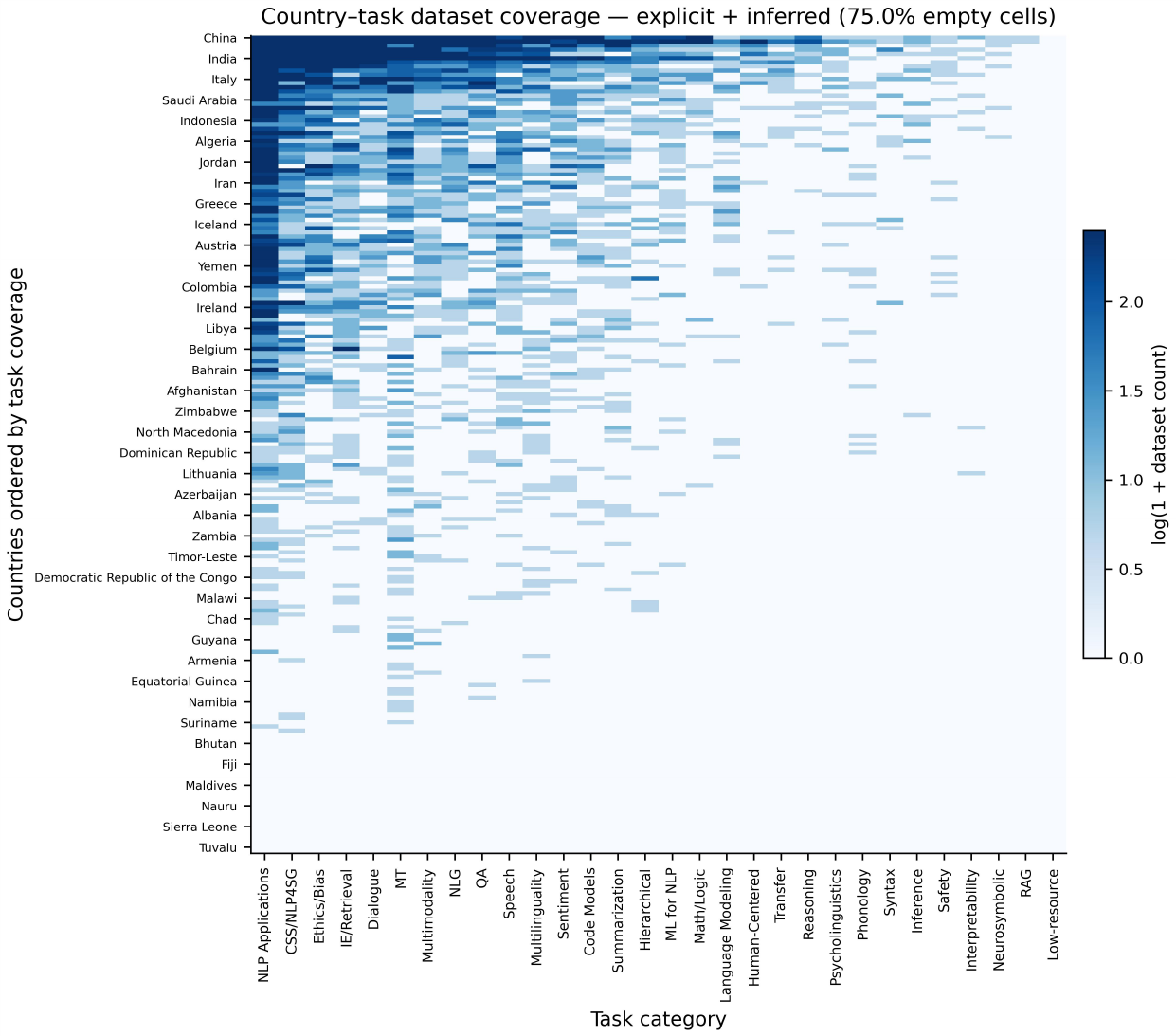}
  \caption{Heatmap of dataset coverage across countries and tasks under explicit+inferred attribution. Coverage is highly fragmented, with most country-task pairs lacking represented dataset records and even high-coverage countries concentrated in a subset of tasks. Only a subset of country labels is shown for readability, with labels placed at regular intervals (approximately every fifth country). Best viewed in color.}
  \label{fig:app_country_task_heatmap}
\end{figure*}

\paragraph{Analytical units and low-count handling.}
Table~\ref{tab:analysis_denominators} summarizes the principal analytical units used in the paper. Country-level coverage counts expanded country-record associations, so a record explicitly representing $N$ countries contributes one association to each. Production-representation analyses additionally expand records over producer countries. Country-task sparsity instead uses the fixed $197\times30$ country-task space.

\begin{table}[H]
\centering
\resizebox{0.98\columnwidth}{!}{
\begin{tabular}{l l r}
\toprule
\textbf{Analysis set} & \textbf{Unit} & \textbf{N} \\
\midrule
AtlasNLP-Core & dataset records & 13,462 \\
Explicit rep. & records / country-record assoc. & 2,447 / 4,421 \\
Explicit+inferred rep. & records / country-record assoc. & 3,506 / 6,001 \\
Production-rep. (explicit) & records w/producer / expanded assoc. & 2,396 / 6,533 \\
Country-task matrix & country-task cells & 5,910 ($197\times30$) \\
\bottomrule
\end{tabular}
}
\caption{Denominator map for the principal AtlasNLP-Core geographic analyses. Different analyses expand paper-level dataset records according to the geographic relation being measured.}
\label{tab:analysis_denominators}
\end{table} 

For production-representation comparisons, the primary summary restricts interpretation to countries with at least 10 represented records with producer metadata and 10 producer-representation associations. Lower-volume countries remain visible in the figure but are treated descriptively because their ratios can change substantially with only a few records. Section~\ref{app:attribution_sensitivity} additionally reports results without this minimum-count restriction and under explicit+inferred attribution.

\subsection{Recovery Diagnostic}
\label{app:temporal_recovery}

Because AtlasNLP-Core is constructed through ACL metadata retrieval, NLI screening, automated extraction, and a subsequent dataset-role audit, differences in pipeline recovery can affect analyses over publication time. We therefore treat temporal patterns as descriptive and use AtlasNLP-Gold only as a reference-set diagnostic rather than as an estimate of population-wide recall.

Among 339 unique ACL-linked Gold papers in the final curated collection, 299 (88.2\%) were recovered in the initial 18,035-record extraction set, and 267 (78.8\%) are retained in the final 13,462-record Core. Conditional on initial recovery, 267/299 (89.3\%) survive the final dataset-role and provenance audit. Of the 32 initially recovered Gold papers subsequently excluded, 28 were classified as reuse of an existing dataset and four as cases in which the target dataset contribution could not be established.

\begin{table}[H]
\centering
\resizebox{0.90\columnwidth}{!}{
\begin{tabular}{lrr}
\toprule
\textbf{Stage} & \textbf{Gold papers recovered} & \textbf{Rate} \\
\midrule
ACL-linked Gold reference & 339 & -- \\
Initial extraction & 299 & 88.2\% \\
Final AtlasNLP-Core & 267 & 78.8\% \\
Final, conditional on initial recovery & 267 / 299 & 89.3\% \\
\bottomrule
\end{tabular}
}
\caption{Reference-set recovery for ACL-linked AtlasNLP-Gold papers across the extraction and final audit stages. Rates are diagnostic of this curated reference set and should not be interpreted as population-wide ACL recall.}
\label{tab:temporal_recovery}
\end{table}
Diagnostic checks conducted during pipeline development also showed that recovery varies across publication periods rather than following a uniform or monotonic pattern. Because both document availability and extraction behavior can vary with publication year, we do not interpret differences in the temporal composition of AtlasNLP-Core as evidence of changes in the underlying NLP dataset ecosystem. Accordingly, the camera-ready analysis does not make substantive claims about temporal diversification of country task portfolios.

\end{document}